\documentclass{article} % For LaTeX2e
\usepackage{natbib}
\usepackage{amsmath,amsfonts,bm}

\def\eqref#1{equation~\ref{#1}}
\def\1{\bm{1}}

\DeclareMathAlphabet{\mathsfit}{\encodingdefault}{\sfdefault}{m}{sl}
\SetMathAlphabet{\mathsfit}{bold}{\encodingdefault}{\sfdefault}{bx}{n}

\newcommand{\R}{\mathbb{R}}

\DeclareMathOperator*{\argmin}{arg\,min}

\usepackage{IRM}
\usepackage{hyperref}
\usepackage{url}
\usepackage{array}
\usepackage{caption}
\usepackage{subcaption}
\usepackage{wrapfig}
\usepackage{floatrow}

\usepackage[top=2cm, bottom=2.5cm, left=2.5cm, right=2.5cm]{geometry}
\title{Lipschitz-Bounded Equilibrium Networks}
\author{Max Revay, Ruigang Wang \& Ian R. Manchester  \\\,\\
Sydney Institute for Robotics and Intelligent Systems (SIRIS)\\
Australian Centre for Field Robotics (ACFR)\\
University of Sydney, Australia \\
\texttt{\{max.revay,ruigang.wang,ian.manchester\}@sydney.edu.au} \\
}

\begin{document}

\maketitle

\begin{abstract}
This paper introduces new parameterizations of equilibrium neural networks, i.e. networks defined by implicit equations. This model class includes standard multilayer and residual networks as special cases. The new parameterization admits a Lipschitz bound during training via unconstrained optimization: no projections or barrier functions are required. Lipschitz bounds are a common proxy for robustness and appear in many generalization bounds. Furthermore, compared to previous works we show well-posedness (existence of solutions) under less restrictive conditions on the network weights and more natural assumptions on the activation functions: that they are monotone and slope restricted. These results are proved by establishing novel connections with convex optimization, operator splitting on non-Euclidean spaces, and contracting neural ODEs. In image classification experiments  we show that the Lipschitz bounds are very accurate and improve robustness to adversarial attacks.
\end{abstract}

% \IM{Title: as above, or just "Robust Equilibrium Networks" if we are focusing on Lipschitz bounds?}

% (Title?) Two Perspectives on Robust Equilibrium Networks: Operator Splitting and Contracting Neural ODE

% Ian's comment: this sounds like just a commentary, not proposing any new model class/method

% Well-Posed and Robust Equilibrium Neural Networks

% On Monotone Operator Equilibrium Networks: Flexible Parameterizations, Lipschitz Bounds, and Connections to Contracting Neural ODEs

% Ray's comment: a little bit long. How about to remove ``monotone operator'' and ``neural ODE'', e.g. Equilibrium Networks with Flexible Parameterizations and Lipschitz Bounds?

% \RW{Ray: since we do not have time to explore the benefits of having the Lyapunov matrix $P$ (various numerical methods and the parameterization with $A$ and $B$), maybe we can leave it for the RNN-TAC paper.}
% \IM{Yes, agree about the $A,B$ parameterisation. I think we can still mention that the function is monotone with respect to $P$}

\section{Introduction}

Deep neural network models have revolutionized the field of machine learning: their accuracy on practical tasks such as image classification  and their scalability have led to an enormous volume of research on different model structures and their properties \citep{lecun2015deep}. In particular, deep residual networks with skip connections \cite{he2016deep} have had a major impact, and neural ODEs have been proposed as an analog with ``implicit depth'' \citep{chen2018neural}. Recently, a new structure has gained interest: \textit{equilibrium networks} \citep{bai2019deep, winstonMonotoneOperatorEquilibrium2020}, a.k.a. \textit{implicit deep learning models} \citep{ghaoui2019implicit}, in which model outputs are defined by implicit equations incorporating neural networks. This model class is very flexible: it is easy to show that includes many previous structures as special cases, including standard multi-layer networks, residual networks, and  (in a certain sense) neural ODEs.

However model flexibility in machine learning is always in tension with model \textit{regularity} or \textit{robustness}. While deep learning models have exhibited impressive generalisation performance in many contexts it has also been observed that they can be very brittle, especially when targeted with adversarial attacks \citep{szegedy2013intriguing}. In response to this, there has been a major research effort to understand and certify robustness properties of deep neural networks, e.g. \cite{raghunathan2018certified, tjeng2018evaluating, liu2019algorithms,cohen2019certified} and many others. Global Lipschitz bounds (a.k.a. incremental gain bounds) provide a somewhat crude but nevertheless highly useful proxy for robustness \citep{tsuzuku2018lipschitz, fazlyab2019efficient}, and appear in several analyses of generalization (e.g.  \citep{Bartlett:2017, Zhou:2019}). 

Inspired by both of these lines of research, in this paper we propose new parameterizations of equilibrium networks with guaranteed Lipschitz bounds. We build directly on the monotone operator framework of \cite{winstonMonotoneOperatorEquilibrium2020} and the work of \cite{fazlyab2019efficient} Lipschitz bounds.

The main contribution of our paper is the ability to enforce tight bounds on the Lipschitz constant of an equilibrium network during training with essentially \textit{no extra computational effort}. In addition, we prove existence of solutions with less restrictive conditions on the weight matrix and more natural assumptions on the activation functions via novel connections to convex optimization and contracting dynamical systems. Finally, we show via small-scale image classification experiments that the proposed parameterizations can provide significant improvement in robustness to adversarial attacks with little degradation in nominal accuracy. Furthermore, we observe small gaps between certified Lipschitz upper bounds and observed lower bounds computed via adversarial attack.

\section{Related work}

\paragraph{Equilibrium networks,  Implicit Deep Models, and Well-Posedness.} As mentioned above, it has been recently shown that many existing network architectures can be incorporated into a flexible model set called an equilibrium network \citep{bai2019deep, winstonMonotoneOperatorEquilibrium2020} or implicit deep model \citep{ghaoui2019implicit}. In this unified model set, the network predictions are made not by forward computation of sequential hidden layers, but by finding a solution to an implicit equation involving a single layer of all hidden units. One major question for this type of networks is its well-posedness, i.e. the existence and uniqueness of a solution to the implicit equation for all possible inputs. \cite{ghaoui2019implicit} proposed a computationally verifiable but conservative condition on the spectral norm of hidden unit weight. In \cite{winstonMonotoneOperatorEquilibrium2020}, a less conservative condition was developed based on monotone operator theory. Similar monotonicity constraints were previously used to ensure well-posedness of a different class of implicit models in the context of nonlinear system identification \citep[Theorem 1]{tobenkin2017convex}.  On the question of well-posedness, our contribution is a more flexible model set and more natural assumptions on the activation functions: that they are monotone and slope-restricted.
% \begin{itemize}
%     \item Winston Kolter, Bai et al, El Ghaoui et al
% \end{itemize}

\paragraph{Neural Network Robustness and Lipschitz Bounds.} The Lipschitz constant of a function measures the worst-case sensitivity of the function, i.e. the maximum ``amplification'' of difference in inputs to differences in outputs. The key features of a good Lipschitz bounded learning approach include a tight estimation for Lipschitz constant and a computationally tractable training method with bounds enforced. For deep networks,  \cite{tsuzuku2018lipschitz} proposed a computationally efficient but conservative approach since its Lipschitz constant estimation method is based on composition of estimations for different layers. Similarly, \cite{ghaoui2019implicit} proposed an estimation for equilibrium networks via input-to-state (ISS) stability analysis. \cite{fazlyab2019efficient} estimates for deep networks based on incremental quadratic constraints and semidefinite programming (SDP) were shown to give state-of-the-art results, however this results was limited to analysis of an already-trained network. The SDP test incorporated into training via the alternating direction method of multipliers (ADMM) in \cite{pauli2020training}, however due to the complexity of the SDP the training times recorded were almost 50 times longer than for unconstrained networks. Our approach uses a similar condition to \cite{fazlyab2019efficient} applied to equilibrium networks, however we introduce a novel direct parameterization method that enables learning robust models via unconstrained optimization, removing the need for computationally-expensive projections or barrier terms.
% \begin{itemize}
%     \item Fazlyab et al, RIKEN, Allgower
% \end{itemize}

\section{Problem Formulation and Preliminaries}

\subsection{Problem statement}
We consider the weight-tied network in which $x\in\R^d$ denotes the input, and $z\in\R^{n}$ denotes the hidden units, $y\in\R^p$ denotes the output, given by the following implicit equation
\begin{equation}\label{eq:implicit}
   z=\sigma(Wz+Ux+b_z), \quad y=W_oz+b_y
\end{equation}
where $W\in\R^{n\times n}$, $U\in\R^{n\times d}$, and $W_o\in\R^{p\times n}$ are the hidden unit, input, and output weights, respectively, $b_z\in\R^n$ and $b_y\in\R^p$ are bias terms. The implicit framework includes most current neural network architectures (e.g. deep and residual networks) as special cases. To streamline the presentation we assume that $\sigma:\R\rightarrow\R$ is a single nonlinearity applied elementwise, although our results also apply in the case that each channel has a different activation function, nonlinear or linear.

Equation~(\ref{eq:implicit}) is termed as an equilibrium network since its solutions are equilibrium points of the difference equation $z^{k+1}=\sigma(Wz^k+Ux+b_z)$ or the ODE $\dot{z}(t)=-z(t)+\sigma(Wz(t)+Ux+b_z)$. Our goal is to learn equilibrium networks (\ref{eq:implicit}) possessing the following two properties: 
\begin{itemize}
  \item {\bf Well-posedness:} For every input $x$ and bias $b_z$, \eqref{eq:implicit} admits a unique solution $z$.
  \item {\bf $\gamma$-Lipschitz:}  It has a finite Lipschitz bound of $\g$, i.e., for any input-output pairs $(x_1,y_1)$, $(x_2,y_2)$ we have $\|y_1-y_2\|_2\leq \g\|x_1-x_2\|_2$.
\end{itemize}

%\IM{Should we call this robustness? Or e.g. $\gamma$-Lipschitz?}
%TODO: Should we add here a brief discussion of the fact that (1) includes deep and residual networks as special cases? This is discussed in other papers but should be mentioned

\subsection{Preliminaries}

\paragraph{Monotone operator theory.} 
The theory of monotone operators on Euclidean space (see the survey \cite{ryu2016primer}) has been extensively applied in the development of equilibrium network \citep{winstonMonotoneOperatorEquilibrium2020}. In this paper, we will use the monotone operator theory on non-Euclidean spaces \citep{bauschke2011convex}, in particular, we are interested in a finite-dimensional Hilbert space $\Hs$, which we identify with $\R^n$ equipped with a weighted inner product $\ip{x}{y}_Q:= y^\top Q x $ where $Q\succ 0$. The main benefit is that we can construct a more expressive equilibrium network set. A brief summary or relevant theory can be found in Appendix~\ref{sec:mono-theory}; here we give some definitions that are frequently used throughout the paper. An operator is a set-valued or single-valued function defined by a subset of the space $A\subseteq \Hs\times\Hs $. A function $f:\Hs\rightarrow\R\cup\{\infty\}$ is proper if $f(x)<\infty$ for at least one $x$. The proximal operators of a proper function $f$ is defined as 
\[
  \begin{split}
    \prox_f^\alpha(x):=\{z\in\Hs\mid z=\argmin_{u}\frac{1}{2}\|u-x\|_Q^2+\alpha f(u)\},
  \end{split}  
\]
%\IM{Do we use subdifferentials in the main text? If not, its definition can go in the appendix. Here we should include only what is needed to understand the statements in the main text} 
where $\|x\|_Q:=\sqrt{\ip{x}{x}_Q}$ is the induced norm. For $n=1$, we only consider the case of $Q=1$. An operator $A$ is monotone if $\ip{u-v}{x-y}_Q \geq 0$ and strongly monotone with parameter $m$ if $\ip{u-v}{x-y}_Q \geq m\|x-y\|_Q^2$ for all $(x,u),(y,v)\in A$. 
%The operator splitting problem is about finding a zero in a sum of two operators $A$ and $B$, i.e. find an $x$ such that $ 0\in (A+B)(x)$. 

\paragraph{Dynamical systems theory.} In this paper, we will also treat the solutions of (\ref{eq:implicit}) as equilibrium points of certain dynamical systems $\dot{z}(t)=f(z(t))$. Then, the well-posedness and robustness properties of (\ref{eq:implicit}) can be guaranteed by corresponding properties of the dynamical system's solution set. A central focus in robust and nonlinear control theory for more than 50 years -- and largely unified by the modern theory of integral quadratic constraints \citep{megretski1997system} -- has been on systems which are interconnections of linear mappings and ``simple'' nonlinearities, i.e. those easily bounded in some sense by quadratic functions. Fortuitously, this characteristic is shared with  deep, recurrent, and equilibrium neural networks, a connection that we use heavily in this paper and has previously been exploited by \cite{fazlyab2019efficient, ghaoui2019implicit, revay2020convex} and others. A particular property we are interested in is called {\em contraction} \citep{lohmiller1998contraction}, i.e., any pair of solutions $z_1(t)$ and $z_2(t)$ exponentially converge to each other:
\[ 
  \|z_1(t)-z_2(t)\|\leq \alpha \|z_1(0)-z_2(0)\| e^{-\beta t}
\]
for all $t>0$ and some $\alpha,\beta>0$. Contraction can be established by  finding a Riemannian metric with respect to which nearby trajectories converge, which is a differential analog of a Lyapunov function. A nice property of a contracting dynamical system is that if it is time-invariant, a unique equilibrium exists and possess certain level of robustness. Moreover, contraction can also be linked to monotone operators, i.e. a system is contracting w.r.t. to a constant (state-independent) metric $Q$ if and only if the operator $-f$ is strongly monotone w.r.t. $Q$-weighted inner product.  We collect some directly relevant results from systems theory in Appendix~\ref{sec:dyn-sys}.

%Integral quadratic constraints (IQCs, \cite{megretski1997system}) provide a powerful framework for robustness analysis of Lur\'{e} systems.

\section{Main Results}

This section contains the main theoretical results of the paper: conditions implying well-posedness and Lipschitz-boundedness of equilibrium networks, and direct (unconstrained) parameterizations such that these conditions are automatically satisfied.

\begin{ass}\label{ass:sigma}
  The activation function $\sigma$ is monotone and slope-restricted in $[0,1]$, i.e.,
\begin{equation}\label{eq:sector_sc}
      0\leq \frac{\sigma(x)-\sigma(y)}{x-y}\leq 1,\;\forall x,y\in\R,\ x\neq y.
\end{equation}

\end{ass}

\begin{remark}
	We will show below (Proposition~\ref{prop:sigma} in Section~\ref{sec:mo}) that Assumption~\ref{ass:sigma} is  equivalent to the assumption on $\sigma$ in \cite{winstonMonotoneOperatorEquilibrium2020}, i.e. that $\sigma(\cdot)=\prox_f^1(\cdot)$ for some proper convex function $f$. However, the above assumption is arguably more natural, since it is easily verified for standard activation functions. Note also that if different channels have different activation functions, then we simply require that they all satisfy (\ref{eq:sector_sc}).
\end{remark}

%\IM{The next two are not really assumptions (that we require to hold), they are conditions that we will impose during the search.}

The following conditions are central to our results on well-posedness and Lipschitz bounds:
\begin{cond}\label{cond:W}
There exists a $\Lambda\in \mathbb{D}^+$ such that $W$ satisfies
\begin{equation}\label{eq:W}
  2\Lambda - \Lambda W-W^T\Lambda\succ 0.
\end{equation}
\end{cond}

\begin{cond}\label{cond:lipschitz}
  Given a prescribed Lipschitz bound $\g>0$, there exists $\Lambda\in \mathbb{D}^+$, with $\mathbb{D}^+$ denoting diagonal positive-definite matrices, such that $W,W_o,U$ satisfy 
  \begin{equation}\label{eq:lipschitz} 
    2\Lambda-\Lambda W-W^T\Lambda-\frac{1}{\gamma}W_o^TW_o-\frac{1}{\gamma}\Lambda U U^T\Lambda\succ 0.
  \end{equation}
\end{cond}
%\IM{To be consistent, we should probably write $\succ 0$ for both of the above. It is equivalent to $\succ \epsilon 0$, since they are just finite-dimensional matrices, but simpler. Also, we can note that (3) implies (2) since the added terms are negative semidefinite}

\begin{remark} Note that Condition~\ref{cond:lipschitz} implies Condition~\ref{cond:W} since $ 1/\g(W_o^TW_o+\Lambda U U^T\Lambda)\succeq 0$. As a partial converse, if Condition~\ref{cond:W} holds, then for any $W_o, U$ there exist a sufficiently large $\g$ such that Condition~\ref{cond:lipschitz} is satisfied. 
\end{remark}

%\IM{This is a property of the condition, not the particular parameterization here. It should go up after (2), (3)}

The main theoretical results of this paper are the following: 
 
\begin{thm}\label{thm:well-posedness}
  If Assumption \ref{ass:sigma} and Condition \ref{cond:W} hold, then the equilibrium network  (\ref{eq:implicit}) is well-posed, i.e. for all $x$ and $b_z$, equation (\ref{eq:implicit}) admits a unique solution $z$. Moreover, it has a finite Lipschitz bound from $x$ to $y$.
\end{thm}

\begin{thm}\label{thm:robustness}
  If Assumption \ref{ass:sigma} and Condition \ref{cond:lipschitz} hold, then the equilibrium network (\ref{eq:implicit}) is well-posed and has a Lipschitz bound  of $\g$.  
\end{thm}

 \begin{wrapfigure}{r}{0.5\linewidth}
	\centering
	\vspace{-8pt}
    \includegraphics[width=0.6\columnwidth]{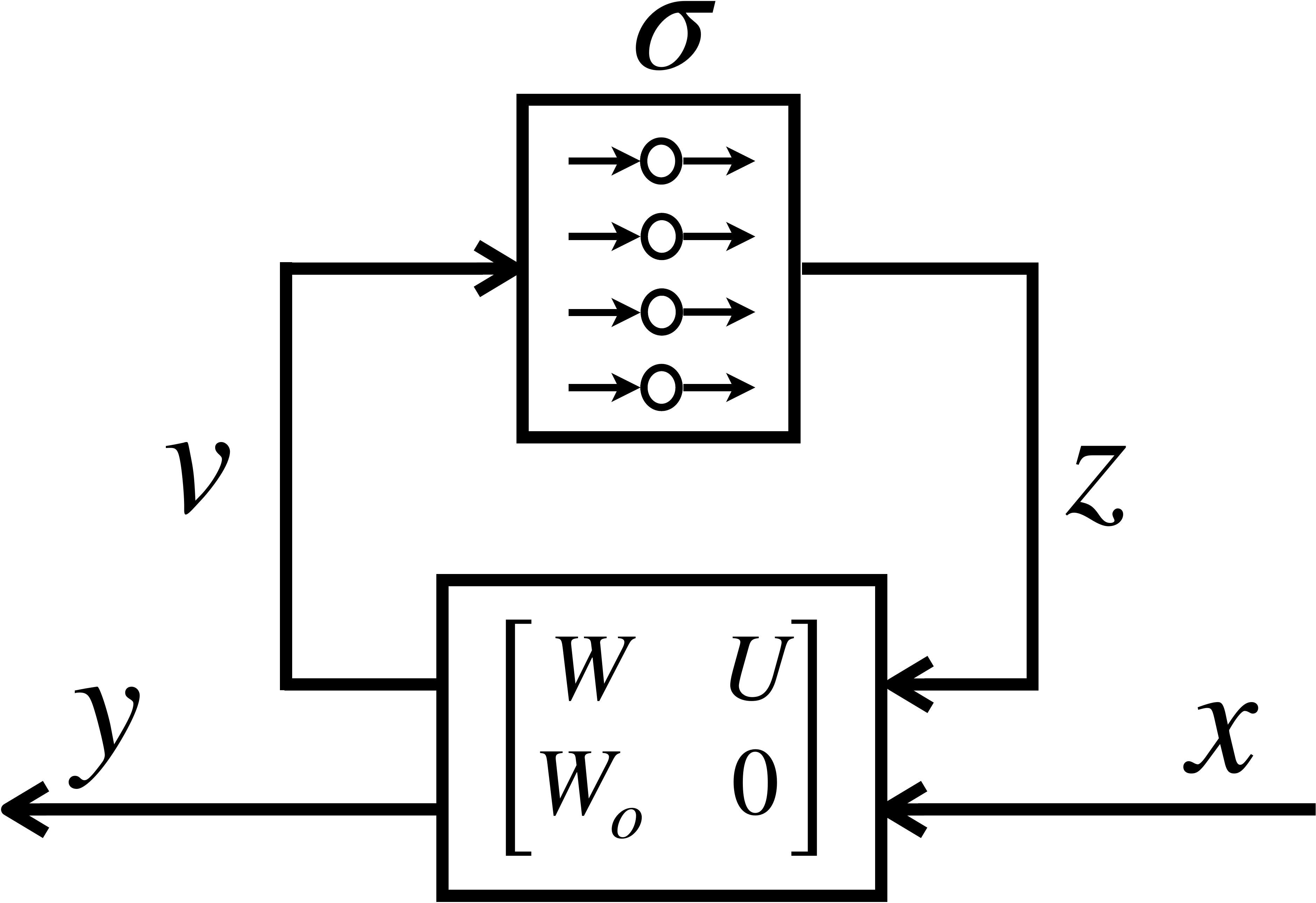}
    \caption{Equilibrium network as a feedback interconnection of a linear mapping and nonlinear activation functions. \vspace{0pt}}\label{fig:lben}
\end{wrapfigure}
%
%\begin{figure}
%  \centering
%    \includegraphics[width=0.3\columnwidth]{figs/fig-lben}
%    \caption{Equilibrium network as a feedback interconnection of a linear mapping and nonlinear activation functions.}\label{fig:lben}
%\end{figure}
As a consequence, we call (\ref{eq:implicit}) a \textit{Lipschitz bounded equilibrium network} (LBEN) if its weights satisfy either (\ref{eq:W}) or (\ref{eq:lipschitz}). We will prove these theorems below, but first we make some straightforward remarks. As depicted in Figure~\ref{fig:lben}, we can represent (\ref{eq:implicit}) by the algebraic feedback interconnection:
\begin{align}\label{eq:affine}
  v = Wz+Ux+b_z,\quad z = \sigma(v),\\ \quad y=W_oz+b_y.\notag
\end{align}
% \begin{align}
%   &v = Wz+Ux+b_z,\label{eq:affine}\\
%   &z = \sigma(v),\label{eq:activation}\\
%   &y = W_oz+b_y.\label{eq:output}
% \end{align}
%\IM{We could put these on one line if we are desperate for space}
% Under Assumption \ref{ass:sigma}, any pair of solutions $z_a=\sigma(v_a), z_b=\sigma(v_b)$ of (\ref{eq:activation}) satisfy \IM{I think for this section, keeping the inner product notation I had before is much more consistent and easy to follow}
% \begin{equation}\label{eq:incr_sector}
%   \begin{bmatrix}
%     \Delta_v \\ \Delta_z
%   \end{bmatrix}^\top
%   \begin{bmatrix}
%     0 & \Lambda \\ \Lambda & -2\Lambda
%   \end{bmatrix}
%   \begin{bmatrix}
%     \Delta_v \\ \Delta_z
%   \end{bmatrix}\geq 0 \quad \forall\Lambda\in \mathbb{D}^+,
% \end{equation}
% where $\Delta_v = v_a-v_b$ and $\Delta_z = z_a-z_b$.
Now, for each activation function, \eqref{eq:sector_sc} can be rewritten as $(x-y)(\sigma(x)-\sigma(y)) \ge (\sigma(x)-\sigma(y))^2$. Clearly any conic (non-negative) combinations of this inequality applied to the individual activations is also true, i.e.  $\sigma$ satisfies the incremental sector condition
\begin{equation}\label{eq:sector}
(v_a-v_b)^T\Lambda (z_a-z_b) \ge  (z_a-z_b)^T\Lambda  (z_a-z_b) 
\end{equation}
for any pair of solutions $z_a=\sigma(v_a), z_b=\sigma(v_b)$ and any $\Lambda\in\mathbb D^+$.
Now, let $\Delta_v = v_a-v_b$ and $\Delta_z = z_a-z_b$, then the sector condition (\ref{eq:sector}) can be rewritten as:
\begin{equation}\label{eq:incr_sector}
  \langle \Delta_v-\Delta_z, \Delta_z\rangle_\Lambda \ge 0.  
\end{equation}
On the other size, the relation (\ref{eq:W}) states that pairs of solutions of (\ref{eq:affine}) with the same input $x$ satisfy
\begin{equation}\label{eq:incr_D}
  \langle\Delta_v-\Delta_z,\Delta_z\rangle_\Lambda \le  -\epsilon |\Delta_z|_\Lambda^2
\end{equation}
for some $\epsilon>0$.  From these it follows that if a solution exists to (\ref{eq:affine}) then it is unique:  (\ref{eq:incr_sector}) and (\ref{eq:incr_D}) taken together imply that $\epsilon \|\Delta_z\|_\Lambda \le 0$ where $\epsilon>0$ and $\Lambda\in\mathbb D^+$, i.e. $\Delta_z=0$. The existence of a solution will be proven via different perspectives in Sections~\ref{sec:mo} and \ref{sec:neural-ode}.

We can also sketch a proof of Theorem~\ref{thm:robustness}. Since Condition~\ref{cond:lipschitz} implies Condition~\ref{cond:W}, from Theorem~\ref{thm:well-posedness} the solutions $z$ exist for all input $x$. For any pair of inputs $x_a$ and $x_b$, let $(v_a,z_a,y_a)$ and $(v_b,v_b,y_b)$ be the solutions to (\ref{eq:affine}), respectively. Their differences satisfy $\Delta_v=W\Delta_z+U\Delta_x$ and $\Delta_y=W_o\Delta_z$. 
%Note that (\ref{eq:incr_sector}) still holds due to Assumption~\ref{ass:sigma} but (\ref{eq:incr_D}) may not be satisfied due to the different inputs. 
To obtain the Lipschitz bound, we first apply Schur complement to (\ref{eq:lipschitz}), then left-multiply by $\begin{bmatrix}
  \Delta_z^\top & \Delta_x^\top 
\end{bmatrix}$ and right-multiply by $\begin{bmatrix}
  \Delta_z^\top & \Delta_x^\top
\end{bmatrix}^\top $, yielding the following:
\[
  \gamma \|\Delta_x\|_2^2-\frac{1}{\gamma}\|\Delta_y\|_2^2\ge 2  \langle\Delta_v-\Delta_z,\Delta_z\rangle_{\Lambda}\geq 0,
\]
where the inequality comes (\ref{eq:incr_sector}). It directly follows that $\|\Delta_y\|_2\le \gamma \|\Delta_x\|_2$ so the network has a Lipschitz bound of $\gamma$. See Appendix~\ref{sec:proof-thm-2} for a detailed proof.

% \IM{We need to actually prove Theorem 2 somewhere. If we keep the above discussion (I think we should) then the proof is easy to write, given that solutions exist for all $x$. I.e. by Schur complement (3) becomes
% \[
% \gamma |\Delta_w|^2-\frac{1}{\gamma}|\Delta_y|^2\ge 2  \langle \Delta_z, \Delta_v-\Delta_z\rangle_{\Lambda}
% \]
% and by \eqref{eq:incr_sector} the right-hand-side is non-negative, so we get the gain bound easily.
% }

%\IM{(cite Glover, D'Amato, Safanov)}
  
\begin{remark} 
  In \cite{fazlyab2019efficient} it was claimed that  (\ref{eq:incr_sector}) holds with a richer (more powerful) class of multipliers $\Lambda$ previously introduced for robust stability analysis of systems with repeated nonlinearities, e.g. recurrent neural networks \citep{chu1999bounds,damato2001new,kulkarni2002all}. However this is not true: a counterexample was given in \cite{pauli2020training}, and here we provide a brief explanation: even if the nonlinearities $\sigma(v_i)$ are repeated when considered as functions of $v_i$, their increments $\Delta_{zi}=\sigma(v_i+\Delta_{vi})-\sigma(v_i)$ are not repeated when considered as functions of $\Delta_{vi}$, since they depend on the particular  $v_i$ which generally differs between units.
 \end{remark}

\begin{figure}
\floatbox[{\capbeside\thisfloatsetup{capbesideposition={right,top},capbesidewidth=9cm}}]{figure}[\FBwidth]
{\caption{Valid coefficient ranges for Example \ref{ex:param}. \vspace{4pt}\\\textit{Gray region:} the condition from \cite{winstonMonotoneOperatorEquilibrium2020} is feasible: $2I-W-W^T\succ 0.$\vspace{4pt}\\\textit{White region (including gray region):} our well-posedness condition is feasible: $\exists \Lambda\in\mathbb D^+ : 2\Lambda-\Lambda W-W^T\Lambda \succ 0.$\vspace{4pt}\\ \textit{Black region:} neither condition feasible. }\label{fig:two_by_two}}
{\includegraphics[width=4.5cm]{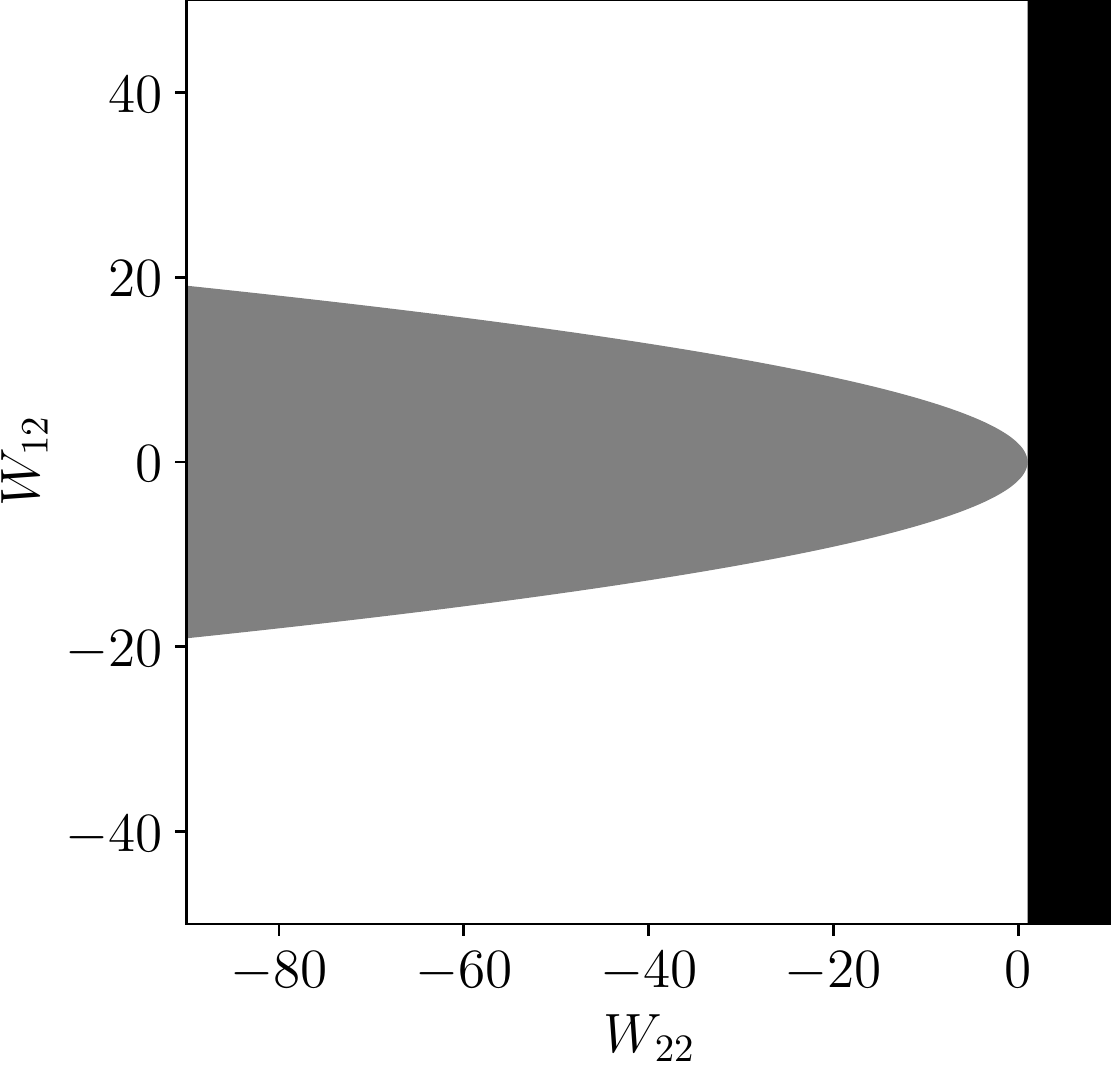}}
\end{figure}

%\RW{To get more space, we could keep Figure 2 but move the analysis of Example 1 to the appendix.}
\begin{ex}\label{ex:param}
We illustrate the extra flexibility of Condition \ref{cond:W} compared to the condition of \cite{winstonMonotoneOperatorEquilibrium2020} by a toy example. 
  Consider $W\in \R^{2\times 2}$ and take a slice near $W=0$ of the form 
  \begin{equation}\label{eq:two_by_two}
    W = \begin{bmatrix}
      0 & W_{12} \\ 0 & W_{22}
    \end{bmatrix}, 
% \end{equation}
\,\, \textrm{for which we have:}\quad
% \begin{equation}
  2I-W-W^T = 
  \begin{bmatrix}
    2 & -W_{12}\\-W_{12} & 2-2W_{22}
  \end{bmatrix}.
  \end{equation}

  By Sylvester's criterion, this matrix is positive-definite if and only if $W_{22}<1$ and $\det (2I-W-W^T) = 4(1-W_{22})-W_{12}^2>0$, which defines a parabolic region in the $W_{12}, W_{22}$ plane.
   
  Applying our condition (\ref{eq:W}), without loss of generality take $\Lambda = \diag(1, \alpha)$ with $\alpha>0$ and we have
  \[
    2\Lambda-\Lambda W-W^T\Lambda = 
    \begin{bmatrix}
      2 & -W_{12}\\-W_{12} & 2\alpha-2\alpha W_{22}
    \end{bmatrix}.
  \]
  The positivity test is now $W_{22}<1$ and $4\alpha(1-W_{22})-W_{12}^2>0$. For each $W_{12}$ there is sufficiently large $\alpha$ such that the second condition is satisfied, since the first implies $1-W_{22}>0$. Hence the only constraint on $W$ is that $W_{22}<1$, which yields a much larger region in the $W_{12}, W_{22}$ plane (see Figure~\ref{fig:two_by_two}). Interestingly, in this simple example with ReLU activation, the condition $W_{22}<1$ is also a necessary condition for well-posedness \cite[Theorem~2.8]{ghaoui2019implicit}.
  
  % Interestingly, while our condition will always be sufficient for existence of a solution, in this simple example with ReLU activation the condition $W_{22}<1$ is also necessary for a solution to exist for all inputs and biases.  We will prove this by contradiction: suppose a solution exists and that $W_{22}\ge 1$, and let input $x=0$ and the second element of the bias $b_{z2}$ be positive. Now, we have
  % \begin{equation}\notag
  % z_2 = \max(W_{22}z_2+b_{z2},0).
  % \end{equation}
  % By definition of the $\max$ operator, it must be the case that $z_2\ge 0$. Since $W_{22}>0$ and $b_{z2}>0$ we then have $W_{22}z_2+b_{z2}>0$, so the ReLU is active, i.e. $z_2 = W_{22}z_2+b_{z2}$. Rearranging, we have
  % \begin{equation}\notag
  %  (1-W_{22})z_2=b_{z2}
  % \end{equation}
  % But the left hand side of this expression is $\le 0$ since $z_2\ge 0$ and $W_{22}\ge 1$, while the right-hand-side is $>0$ by assumption on the bias. Hence we have a contradiction.
  
\end{ex}
  
%\begin{figure}
%  \centering
%    \includegraphics[width=0.35\columnwidth]{figs/fig-two-by-two}
%    \caption{Valid coefficient ranges for Example \ref{ex:param}. Gray region: condition $2I-W-W^\top\succ 0$ from \cite{winstonMonotoneOperatorEquilibrium2020} is feasible. White region (including gray region): our condition $\exists \Lambda\in\mathbb D^+ : 2\Lambda-\Lambda W-W^\top\Lambda \succ 0$ is feasible. Black region: neither condition feasible.}\label{fig:two_by_two}
%\end{figure}

\subsection{Direct Parameterization for Unconstrained Optimization}

%\RW{section title: Direct Parameterization for removing LMI constraints?}
%\IM{The following text doesn't really make its point clear. I think we should introduce the concept of a "direct parameterization" or "unconstrained parameterization" or something similar, i.e. where the feasible set is just a known mapping from $\R^q$ for some $q$, as opposed to having any constraints imposed.}

Training a network that satisfies Condition~\ref{cond:W} or \ref{cond:lipschitz} can be formulated as a constrained optimization problem. In fact, Condition~\ref{cond:W} is a linear matrix inequality (LMI) in the variables $\Lambda$ and $\Lambda W$, from which $W$ can be determined uniquely. Similarly, via Schur complement, Condition~\ref{cond:lipschitz} is an LMI in the variables $\Lambda, \Lambda W, \Lambda U, W_o,$ and $\gamma$, from which all network weights can be determined. In a certain theoretical sense LMI constraints are tractable -- \cite{nesterov1994interior} proved they are polynomial-time solvable -- however for even for moderate-scale networks (e.g. $\le 100$ activations) the associated barrier terms or projections become a major computational bottleneck,  and for large-scale networks they quickly become prohibitive.

In this paper we propose direct parameterizations that allows learning via  unconstrained optimization problem, i.e. all network parameters are transformations of free (unconstrained) matrix variables, in such a way that LMI constraints (\ref{eq:W}) or (\ref{eq:lipschitz}) are automatically satisfied.

%By choosing arbitrary $\Psi\in\mathbb{D}^+$, $S=-S^T\in\R^{n\times n}$ and $V\in\R^{n\times n}$, we can construct a hidden unit weight 
For Condition (\ref{eq:W}), we parameterize via the following free variables: a matrix $V\in\R^{n\times n}$, a vector $d\in\R^n$, and skew-symmetric\footnote{Note that $S$ can be parameterized via its upper or lower triangular components, or via $S= N-N^T$ with $N$ free, which can be more straightforward if $W$ is defined implicitly via linear operators, e.g. convolutions.} matrix $S=-S^T \in\R^{n\times n}$,  we can construct the hidden unit weight 
\begin{equation}\label{eq:direct_rep}
  W = I-\Psi(V^TV+\epsilon I+S),
\end{equation}
where $\Psi=\diag \left(e^d\right)$ and $\epsilon>0$ is some small constant to ensure strict positive-definiteness. Then it follows from straightforward manipulations that Condition \ref{cond:W} holds with $\Lambda=\Psi^{-1}$ if and only if $W$ can be constructed as in (\ref{eq:direct_rep}). When $\Psi=I$, i.e. $d=0$, this is exactly the parameterization used in \cite{winstonMonotoneOperatorEquilibrium2020}. 

Similarly, for  Condition~\ref{cond:lipschitz}, we add to the parameterization the free input and output weights  $U$ and $W_o$, and arbitrary $\gamma>0$, we can construct
\begin{equation}\label{eq:direct_gamma}
  W = I-\Psi\left(\frac{1}{2\gamma}W_o^TW_o+\frac{1}{2\gamma} \Psi^{-1} U U^T \Psi^{-1}+V^TV+\epsilon I+S\right),
\end{equation} 
for which  (\ref{eq:lipschitz}) is automatically satisfied. Again, it can easily be verified that this construction is necessary and sufficient, i.e. any $W$ satisfying (\ref{eq:lipschitz}) can be constructed via (\ref{eq:direct_gamma}).

%The condition (\ref{eq:W}) includes as a special case the condition of \cite{winstonMonotoneOperatorEquilibrium2020} with $\Lambda=I$. The extra flexibility of our condition is  by the following simple example.

%\IM{Again, this is a property of the condition. However, we can say that with $\Psi=I$ this is exactly the parameterization used by W\&K}

%\begin{wrapfigure}{r}{0.5\linewidth}
%	\centering
%	\vspace{-10pt}
%    \includegraphics[width=0.4\columnwidth]{figs/fig-two-by-two}
%    \caption{Valid coefficient ranges for matrices of the form (\ref{eq:two_by_two}). Gray region: condition $2I-W-W^\top\succ 0$ from \cite{winstonMonotoneOperatorEquilibrium2020} is feasible. White region (including gray region): our condition $\exists \Lambda\in\mathbb D^+ : 2\Lambda-\Lambda W-W^\top\Lambda \succ 0$ is feasible. Black region: neither condition feasible.}\label{fig:two_by_two}
%    \vspace{-10pt}
%\end{wrapfigure}

\subsection{Convex Optimization and Monotone Operator Perspective}\label{sec:mo}

In this section, we will show that the equilibrium network (\ref{eq:implicit}) is an optimality condition for a strongly convex optimization problem, and hence a solution exists. First, we need the following observation on the activation function $\sigma$. 

\begin{prop}\label{prop:sigma}
	Assumption \ref{ass:sigma} holds if and only if there exists a convex proper function $f:\R\rightarrow\R\cup \{\infty\}$ such that $ \sigma(\cdot)=\prox_f^1(\cdot)$. 
\end{prop}
The proof of Proposition \ref{prop:sigma} with a construction of $f$ appears in Appendix~\ref{sec:proof-lem-1}. Here we provide a list of $f$ for popular $\sigma$ in Table~\ref{tab:sigma-f}. It is well-known in monotone operator theory \citep{ryu2016primer} that for any convex closed proper function $f$, the proximal operator $\prox_f^1(x)$ is monotone and non-expansive (i.e. slope-restricted in $[0,1]$). Proposition \ref{prop:sigma} is a converse result for scalar functions.

\begin{remark}
To our knowledge Proposition \ref{prop:sigma} is novel, however for several popular activation functions the corresponding  functions $f$ were computed in \cite{li2019lifted} (see also Table~\ref{tab:sigma-f} in Appendix~\ref{sec:proof-prop-1}). Compared with \cite{li2019lifted}, our work gives a necessary and sufficient conditions.
	% For most activation functions, e.g. ReLU, the relations in Table \ref{tab:sigma-f} were already known and used in  \cite{bibiDeepLayersStochastic2019, winstonMonotoneOperatorEquilibrium2020}, whereas other activations (e.g. Softplus, Tanh, Sigmoid) were related to convex functions with visually similar proximal operators. In contrast, we provide an exact expression for all activation functions satisfying (\ref{eq:sector_sc}).
\end{remark}

%\IM{Mention that other papers have found expressions which are visually similar, but these are exact} 

Now we connect the equilibrium network (\ref{eq:implicit}) to a convex optimization problem.
\begin{prop}\label{prop:operator}
  If Assumption~\ref{ass:sigma} and Condition \ref{cond:W} hold, then finding a solution of the equilibrium network (\ref{eq:implicit}) is equivalent to solving the following strongly convex optimization problem
  \begin{equation}\label{eq:cv_prob}
    \min_{z} \; J(z)=\left\langle \frac{1}{2}(I-W)z-Ux-b_z,z\right\rangle_{\Lambda}+\mathfrak{f}(z).
  \end{equation}
  where $\mathfrak{f}(z):=\sum_{i=1}^n \lambda_{i}f(z_i)$ with $\lambda_i$ as the $i$th diagonal element of $\Lambda$.
\end{prop}
%\IM{Should $b$ be $Ux+b_z$ here?}

The proof appears in Appendix \ref{sec:proof-prop-1} and Theorem~\ref{thm:well-posedness} follows directly since $J(z)$ is a strongly convex function on $\R^n$ and has a unique minimizer, which is also the solution of (\ref{eq:implicit}). 

%Since the proximal operator of $\mathfrak{f}$ is well-defined function $\sigma$, then $\mathfrak{f}$ is bounded from below. Then, Theorem~\ref{thm:well-posedness} follows by the above proposition since the cost function $J(z)$ is bounded from below and its first term is strongly convex.  

%\IM{We still need more here. Strongly convex? $x\mapsto e^{-x}$ is strictly convex and bounded from below but does not have a minimum} 

%\IM{I think solving and backpropagation should be discussed separately, and we should at least mention FISTA, ADMM and its relation to Douglas Rachford.}

\paragraph{Computing an equilibrium.}  The convex optimization problem (\ref{eq:cv_prob}) has a structure that is amenable to various methods based on ``splitting'', since splits into a strongly-convex quadratic term and a summation of scalar convex functions with simple proximal operators. In particular, ADMM \citep{boyd2011distributed} and FISTA \citep{beck2009fast}, and  Peaceman-Rachford splitting \citep{kellogg1969nonlinear} can be directly applied. \cite{winstonMonotoneOperatorEquilibrium2020} found that Peaceman-Rachford splitting converges very rapidly when properly tuned, and our experience agrees with this. As an alternative, FISTA often requires more iteration steps but does not require computing a matrix inverse or solving a linear system, which can be a significant advantage for large-scale networks.

%\RW{further comments on comparison with DR (equivalent to ADMM), PR and FISTA if we have done some numerical tests.}

%\IM{I would say that FISTA and ADMM more directly approach (14), not first transform into operator splitting, although that's a perspective}

%This algorithm requires to compute the proximal $\prox_{\mathfrak{f}}^\alpha(\cdot)$ in each step and the matrix inverse $(I+\alpha (I-W))^{-1}$ once $W$ is updated. For some simple activations, we can construct the analytical representation of $ \prox_{\mathfrak{f}}^\alpha(\cdot)$, e.g. $\prox_{\mathfrak{f}}^\alpha(\cdot)=\sigma(\cdot)$ for all $\alpha>0$ if $\sigma$ is ReLU. For general cases, it may be difficult to compute the exact representation. However, when $\alpha $ is sufficiently small, $\prox_{\mathfrak{f}}^\alpha(\cdot)$ can be approximated by $((1-\alpha)I+\alpha \sigma)(\cdot)$. 
%This approximation technique can also be applied to compute $ (I+\alpha (I-W))^{-1}\approx (1-\alpha)I+\alpha W$, which can reduce the training time for large-scale networks. 

\paragraph{Gradient backpropagation.} As shown in \citep[Section 3.5]{winstonMonotoneOperatorEquilibrium2020}, the gradients of the loss function $\ell(\cdot)$ can be represented by 
\begin{equation}\label{eq:gradient}
  \frac{\partial \ell}{\partial (\cdot)}=\frac{\partial \ell}{\partial z_\star}(I-JW)^{-1}J\frac{\partial (Wz_\star+Ux+b_z)}{\partial (\cdot)}
\end{equation}
where $z_\star$ denotes the solution of (\ref{eq:implicit}), $(\cdot)$ denotes some learnable parameters in the parameterization (\ref{eq:direct_rep}) or (\ref{eq:direct_gamma}), and $J\in\mathrm{D}\sigma(Wz_\star+Ux+b_z)$ with $\mathrm{D}\sigma$ as the Clarke generalized Jacobian of $\sigma$. Since $\sigma$ is piecewise differentiable, then the set $\mathrm{D}\sigma(Wz_\star+Ux+b_z)$ is a singleton almost everywhere. The following proposition reveals that (\ref{eq:gradient}) is well-defined, see proof in Appendix~\ref{sec:proof-prop-2}.
\begin{prop}\label{prop:gradient}
	The matrix $I-J W$ is invertible for all $z_\star$, $x$ and $b_z$.
\end{prop}

%Equation~(\ref{eq:gradient}) can also be efficiently solved by Peaceman-Rachford algorithm \citep{winstonMonotoneOperatorEquilibrium2020}. \IM{What needs to be solved here via PR? Once you have $z_\star$ from equilibrium solving, you just need parameter gradients and a matrix inversion} \RW{Here $J$ is an operator which may have multiple values. So (\ref{eq:gradient}) is a resolvent operator which can solved by PR. Maybe Proposition 3 needs to change to ``The operator $I-JW$ is strictly monotone''. Todo: more explanation is needed.}

\subsection{Contracting neural ODEs}\label{sec:neural-ode}

In this section, we will prove existence of a solution to (\ref{eq:implicit}) from a different perspective: by showing it is the equilibrium of a contracting dynamical system (a ``neural ODE''). We first add a smooth state $v(t)\in\R^n$ to avoid the algebraic loop in (\ref{eq:affine}). This idea has long been recognized as helpful for well-posedness questions \citep{zamesRealizabilityConditionNonlinear1964}. We define the dynamics of $v(t)$ by the following ODE:
\begin{equation}\label{eq:neural-ode}
  \dot v(t) = -v(t)+Wz(t)+Ux+b_z, \quad z(t) = \sigma(v(t)).
\end{equation}
The well-posedness of (\ref{eq:implicit}) is equivalent to the existence and uniqueness of an equilibrium of (\ref{eq:neural-ode}) for all $x$ and $b_z$, which is established by the following proposition. 

%That is, for any pair of feasible solutions $(v_a(t),z_a(t))$ and $(v_b(t),z_b(t))$, their differences $\Delta_v=v_a-v_b$ and $\Delta_z=z_a-z_b$ converge to $0$. 

\begin{prop}\label{prop:contraction}
  If Assumption~\ref{ass:sigma} and Condition~\ref{cond:W} hold, then the neural ODE (\ref{eq:neural-ode}) is contracting w.r.t. some constant metric $P\succ0$.
\end{prop}
The proof is in Appendix~\ref{sec:proof-prop-4}. Moreover, the metric $P$ can be found via semidefinite programming. The above proposition also proves that the nonlinear operator $-f$ with $f(v)=-v+W\sigma(v)+Ux+b_z$, zeros of which define solutions of the equilibrium network (\ref{eq:implicit}), is actually monotone w.r.t. the $P$-weighted inner product, which gives a first-order cutting-plane oracle for the zero location $v_\star$ such that $f(v_\star)=0$. I.e. given a test point $v_t \ne v^\star$, it proves that $v_\star$ is in the half-space defined by
$
  \langle v_\star-v_t, f(v_t)\rangle_P>0.
$
This may offer alternative ways to solve the equilibrium network (\ref{eq:implicit}), e.g. via \cite{nemirovskiProxMethodRateConvergence2004, nesterovDualExtrapolationIts2007}. 

Note also that the contraction property is independent of the input $x$ and biases, and so extends directly to the case when these are time-varying. Roughly speaking: for any well-posed equilibrium network, there corresponds a contracting (strongly stable) neural ODE.

% \[
%   \ip{f(v_b)-f(v_a)}{v_a-v_b}_P\geq \beta \|v_a-v_b\|_P^2.
% \] 

\section{Experiments}

In this section we test our approach on the MNIST and SVHN image classification problems.
Our numerical experiments focus on model robustness, the trade-off between model performance and the Lipschitz constant, and the tightness of the Lipschitz bound. We compare the the proposed LBEN to unconstrained equilibrium networks, monotone operator equilibrium network (MON) of \cite{winstonMonotoneOperatorEquilibrium2020}, and fully connected networks trained using Lipschitz margin training (LMT) \citep{tsuzuku2018lipschitz}. 
When studying model robustness to adversarial attacks, we use the L2 Fast Gradient Sign Method,  implemented as part of the Foolbox toolbox \citep{rauber2017foolboxnative}. All models are trained on a standard desktop computer with an NVIDIA GeForce RTX 2080 graphics card.
Details of the models and training procedure can be found in Appendix \ref{sec:training_details}. 
%all code will be made available online but links are omitted due to the double-blind review process. 

%can be found at \MR{Upload code to cyberspace} \IM{Can we do code links with double-blind?}

%For a model $S:\mathbb{R}^d \mapsto \mathbb{R}^p$, a lower bound on the Lipschitz constant can be calculated by (approximately) solving:
%\begin{equation}
%	\gamma_{\mathrm{lower}} = \max_{u, v} \frac{|S(u) - S(v)|}{|u-v|}.
%\end{equation}
%using gradient descent. \IM{I think (17) is unnecessary use of space, we already define Lipschitz and I think adversarial attacks are well known in this community}

%We demonstrate the benefits of our proposed parametrization on the benchmark MNIST and CIFAR-10 datasets.
%Firstly we will compare our model to the monotone operator equilibrium network. 

%We will then show how the proposed parametrization can be used to train robust models and compare our model with fully connected networks with the same number of parameters trained using Lipschitz margin training. Models will be compared by their robustness to adversarial perturbations.

\subsection{MNIST Experiments with Fully-Connected Networks}

\begin{figure}
\centering
\begin{subfigure}[t]{.47\textwidth}
  \centering
  \includegraphics[width=1\columnwidth, trim={2cm, 0.5cm, 3cm, 1.5cm}, clip]{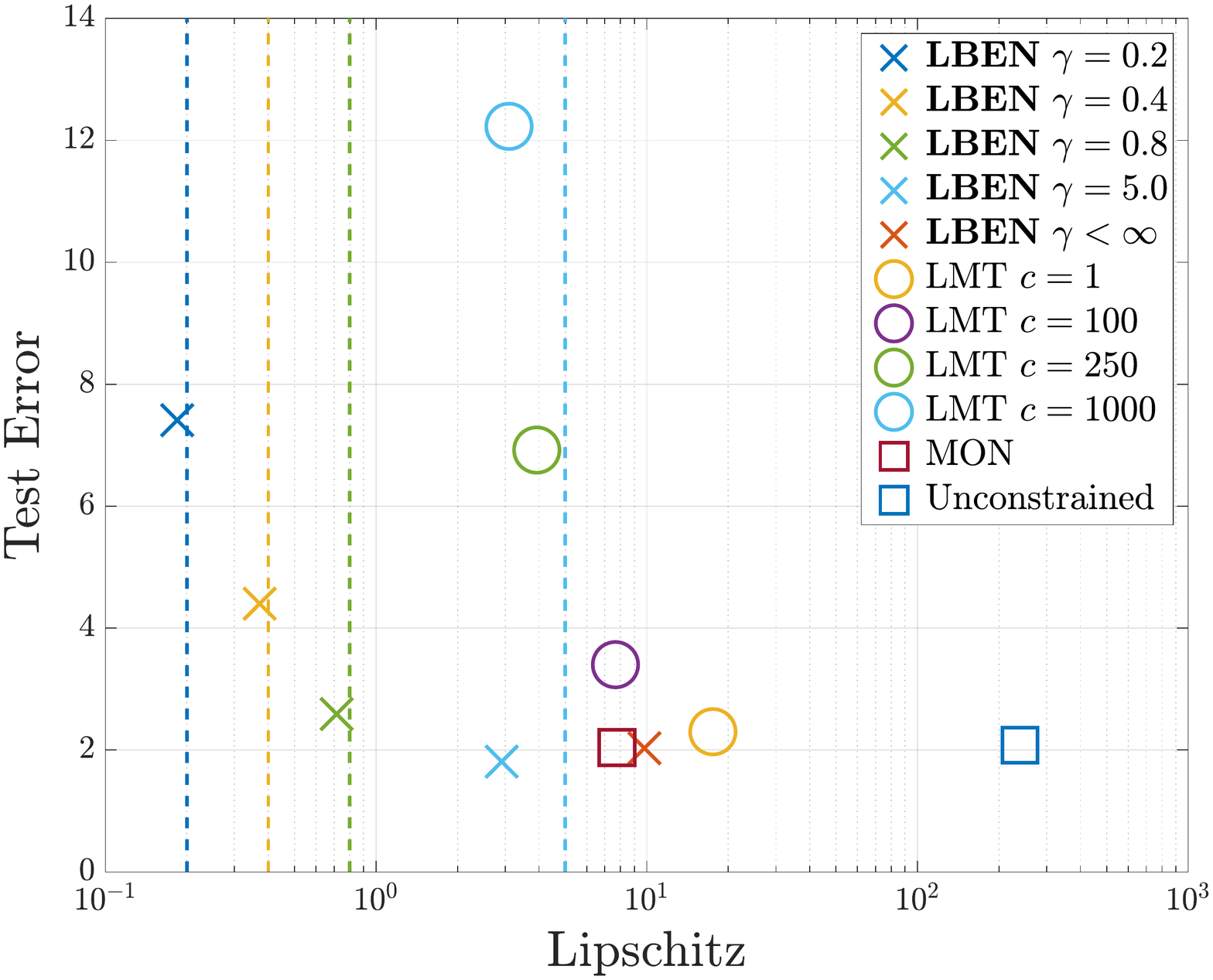}
  \caption{Nominal test error vs Lipschitz constant estimates: markers indicate observed lower bounds for all methods, vertical lines indicate certified upper bounds for LBEN}
  \label{fig:Error_vs_Lipschitz}
  \label{fig:mnist}
\end{subfigure}\hfil
\begin{subfigure}[t]{.47\textwidth}
  \centering
  \includegraphics[width=1\linewidth, trim={2cm, 0.5cm, 3cm, 1.5cm}, clip]{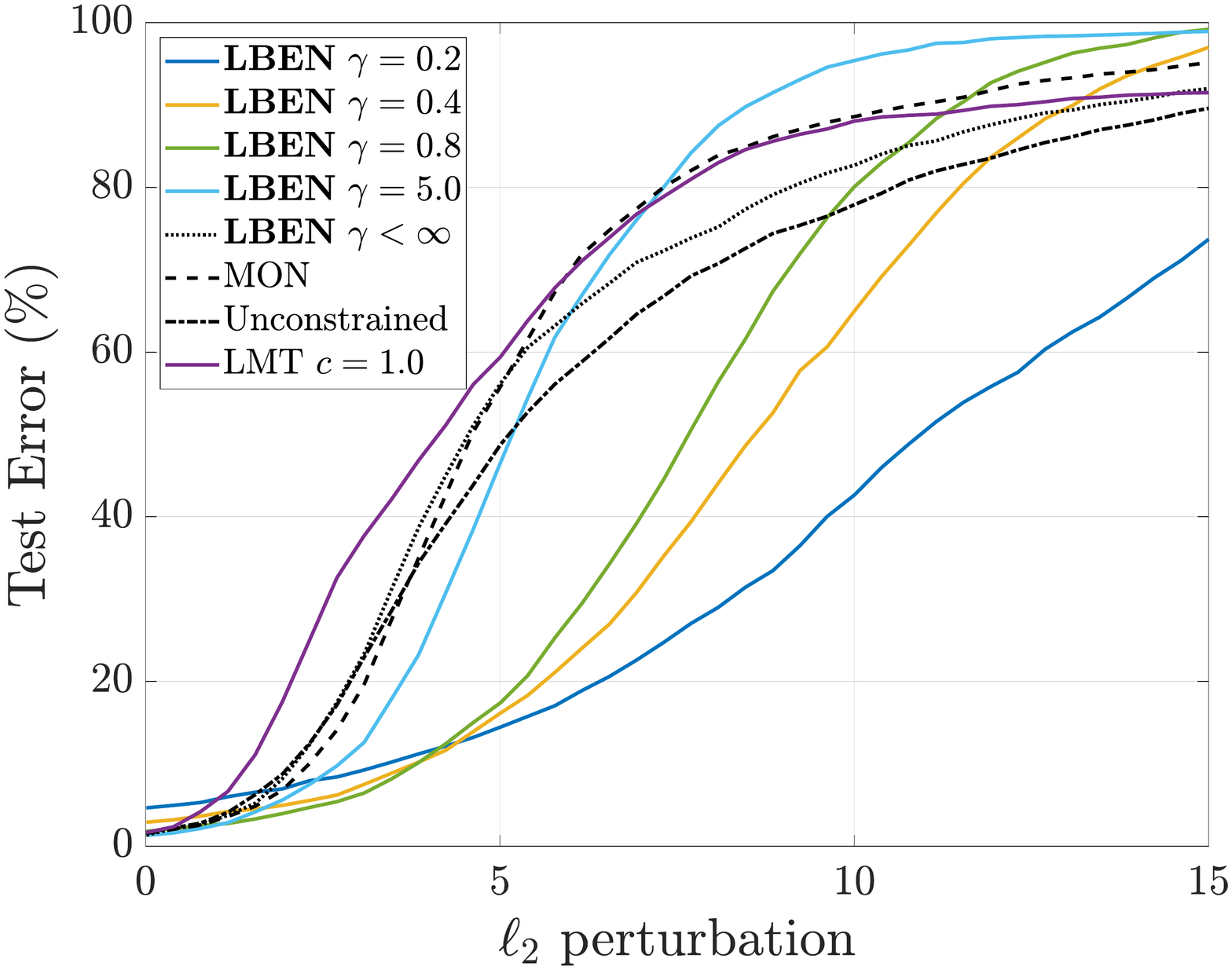}
  \caption{Test error with adversarial perturbation versus size of adversarial perturbation. Lower is better.}
  \label{fig:MNIST_Robustness}
\end{subfigure}

\caption{Image classification results on MNIST character recognition data set.}
\label{fig:test}
\end{figure}

In Figure \ref{fig:mnist} the test error versus the observed Lipschitz constant, computed via adversarial attack for each of the models trained. 
We can see clearly that the parameter $\gamma$ in LBEN offers a trade-off between test error and Lipschitz constant. 
Comparing the $\mathrm{LBEN}_{\gamma=5}$ with both MON and $\mathrm{LBEN}_{\gamma<\infty}$, we also note a slight regularizing effect in the lower test error.

By comparison, LMT \citep{tsuzuku2018lipschitz} with  $c$ as a tunable regularization parameter displays a qualitatively similar trade-off, but underperforms LBEN in terms of both test error and robustness. If we examine the unconstrained equilibrium model, we observe a Lipschitz constant more than an order of magnitude higher, i.e. this model has regions of extremely high sensitivity, without gaining any accuracy in terms of test error. %This supports prior observations that stability constraints \MR{Del?} can improve model generalization and robustness \MR{CN}. By appropriate choice of input the unconstrained equilibrium model can be made unstable and extremely high Lipschitz constant is observed. 
 
For the LBEN models, the lower and upper bounds on the Lipschitz constant are very close: the markers are very close to their corresponding lines in Figure \ref{fig:mnist}, see also the table of numerical results in Appendix \ref{sec:MNIST_stats} in which the approximation accuracy is in many cases around 90\%. %The exception is $\gamma=5$, for which it is likely that the constraint was having a mild effect since the Lipschitz bound for LBEN$_{\gamma<\infty}$ is very similar.

Next we tested robustness of classification accuracy to adversarial attacks of various sizes, the results are shown in Figure \ref{fig:MNIST_Robustness} and summarized in Table \ref{tab:MNIST_results}. We can clearly see that decreasing $\gamma$ (i.e. stronger regularization) in the LBEN models results in a far more gradual degradation of performance as perturbation size increases, with only a mild impact on nominal (zero perturbation) test error.

Finally, we examined the impact of our parameterization on computational complexity compared to other equilibrium models.  The test and training errors versus number of epochs are plotted in Figure \ref{fig:mnist_training}, and we can see that all models converge similarly, and also take roughly the same amount of time per epoch. This is a clear contrast to the results of \cite{pauli2020training} in which imposing Lipschitz constraints resulted in fifty-fold increase in training time. Interestingly, we can also see in Figure \ref{fig:mnist_training} the effect of regularisation for LBEN with $\gamma=5$: higher training error but lower test error.

It should also be noted that we have observed a number of cases where the unconstrained equilibrium model can become unstable during training as solutions are not guaranteed. LBEN never exhibits this problem.

%The complete results for this experiment can be found in Table \ref{tab:MNIST_results}.

\subsection{SVHN Experiments with Convolutional Networks}
The previous example looked at a simple fully connected model, however the approach can also be applied to alternate model structures. We have performed experiments with convolutional equilibrium networks on the SVHN dataset, comparing convolutional LBEN with a convolutional MON. % and show that similar performance can be achieved with reduced Lipschitz constant. %Details of the model structure and training procedure can be found in Appendix \ref{sec:svhn_training_details}.

Table \ref{tab:SVHN_results} in Appendix \ref{sec:MNIST_stats} shows that for a slight decrease in nominal test performance we can reduce the Lipschitz sensitivity to adversarial attack by more than a factor of 10, and significantly increase robustness of classification performance. Note that the Lipschitz bound for this model is not as tight as the one observed in the MNIST example, perhaps because we have used a slightly more restrictive set of multipliers (c.f. Section \ref{sec:svhn_training_details} for details). Further exploration of larger and more richly-structured convolutional networks is a topic of our on-going research.

%\begin{table} \label{tab:svhn}
%	\centering
%	\begin{tabular}{|c|c|c|c|}
%	\hline	Model 					& $\mathrm{LBEN}_{\gamma=2}$ &   MON   \\	\hline \hline	
%	Test Error (\%)			 		& 22.75 &    19.5 \\ \hline
%	Adversarial Error ($\ell_2< 5$) & 37	&    40 		 \\ \hline
%	Adversarial Error ($\ell_2< 10$)& 56  	&    72 	 \\ \hline\hline	
%	$\gamma_\mathrm{upper}$ 		& 2 	&     -	 \\	\hline		
%		$\gamma_\mathrm{lower}$ 	& 0.8 	&    8.3 	 \\	\hline		
%	\end{tabular}
%\caption{Tightness of Lipschitz bounds for LBEN: Test error, certified upper bounds, observed lower bounds, and their ratio for convolutional networks on the SVHN dataset.}
%\end{table}

%
%\begin{figure}
%	\centering
%	\includegraphics[width=0.9\linewidth]{figs/mnist_robustness}
%	\caption{Test error on MNIST with adversarial perturbation versus size of adversarial perturbation. Lower is better.}
%	\label{fig:MNIST_Robustness}
%\end{figure}
%\begin{figure}
%	\centering
%	\includegraphics[width=0.9\linewidth]{figs/Error_vs_Lipschitz}
%	\caption{Test error on MNIST with adversarial perturbation versus size of adversarial perturbation. Lower and further to the left is better.}
%	\label{fig:Error_vs_Lipschitz}
%\end{figure}

\section{Conclusions}
In this paper we have shown that the flexible framework of equilibrium networks can be made robust via a simple and direct parameterization which results in guaranteed Lipschitz bounds. Although we have not explored it in detail in this paper, our results can also be directly applied (as a special case) to standard multilayer and residual deep neural networks, and also provide a direct parameterization of nonlinear ODEs satisfying strong stability and robustness properties. Furthermore, although in this paper we have limited attention to standard scalar activation functions such as ReLU or sigmoids, our results easily extend to certain multivariable ``activations'' that satisfy appropriate monotonicity properties, or more generally integral quadratic constraints. This includes, for example, computing the $\arg\min$ of a quadratic program of the sort that appears in constrained model predictive control \citep{heath2007zames}. Exploring these variations will be a topic of our future research.

\bibliographystyle{plainnat}

\bibliography{RNN, iclr2021_conference}

\clearpage
\newpage

\appendix

\section{Experimental Results on MNIST Character Recognition} \label{sec:MNIST_stats}

%	\begin{tabular}{|c|c|c|c|c|c|c|c|c|}
%				\hline Model&  \multicolumn{7}{c|}{\textbf{LBEN}} & Lip-NN \footnotemark\\ \hline\hline
%				Test Error (\%) 			& 7.41	& 5.44 	& 4.44 	& 3.52	& 2.59	 & 2.36	& 1.81 & 3.55\\ \hline
%				\hline $\gamma$: certified upper bound & 0.2 	& 0.3 	& 0.4 	& 0.5 	& 0.8    & 1.0	& 5.0 & 8.74 \\ \hline
%		       $\gamma$: observed lower bound & 0.184 & 0.282 & 0.372 & 0.458 & 0.715	 & 0.865	& 2.912& - \\ \hline
%		      Ratio lower / upper 		& 0.92 	& 0.94 	& 0.93 	& 0.916 & 0.8937 & 0.865	& 0.5824& -\\ \hline
%	\end{tabular}

This appendix contains tables of results on MNIST and SVHN data sets.

Legend: 
\begin{itemize}
  \item Err: Test error (\%),
  \item $\|a\|_2$: $\ell^2$ norm of adversarial attack.
  \item $\gamma_{up}$: certified upper bound on Lipschitz constant (for models that provide one).
  \item $\gamma_{low}$: observed lower bound on Lipschitz constant via adversarial attack.
  \item $\gamma$ approx: approximation ratio of Lipschitz constant as percentage =  $100\times\left(\frac{\gamma_{low}}{\gamma_{up}}\right)$. 
\end{itemize}
Models:
\begin{itemize}
\item LBEN: the proposed Lipschitz bounded equilibrium network..
\item MON: the monotone operator equilibrium network of \cite{winstonMonotoneOperatorEquilibrium2020}.
\item UNC: an unconstrained equilibrium network, i.e. $W$ directly parameterized.
\item LMT: Lipschitz Margin Training model as in \cite{tsuzuku2018lipschitz}.
\item Lip-NN: The Lipschitz Neural Network model of \cite{pauli2020training}. Note these figures are as reported in \citep{pauli2020training}, all other figures are calculated by the authors of the present paper.
\end{itemize}

\begin{table}[h]
\centering
\begin{tabular}{|c||c|c|c|c|c|c|}\hline
Model& Err:  $\|a\|_2 = 0$ & Err:  $\|a\|_2 \le 5$ &Err:  $\|a\|_2 \le  10$ & $\gamma_{up}$ & $\gamma_{low}$ & $\gamma$ approx
\\ \hline\hline
LBEN$_{\gamma<\infty}$& 2.03 & 56.0  &82 & - 	& 9.8 & -
\\ \hline
LBEN$_{\gamma=5}$	&\textbf{1.81} & 46.4 & 95.4		&5		&2.912	& 58.2\%
\\ \hline
LBEN$_{\gamma=1}$	&	2.36 & 19.4 & 85.5		&1		&0.865	& 86.5\%
\\ \hline
LBEN$_{\gamma=0.8}$	&	2.59 & 17.4	& 80.1		&0.8	&0.715	& 89.4\%
\\ \hline
LBEN$_{\gamma=0.4}$	&	4.44 & 16.1	& 65.0		&0.4	&0.372	& 93\%
\\ \hline	
LBEN$_{\gamma=0.2}$	&	7.41 & \textbf{14.4}	& \textbf{42.6}	&0.2	&0.184	& 92\%
\\ \hline
MON					&   2.04 & 55.8 & 88.6 		& - 	& 7.75 & -
\\ \hline
UNC					&	2.08 & 48.75 & 77.9 	& - 	& 239.0 & -
\\ \hline
LMT$_{c=1}$			&   2.3  & 59.4 & 88.1 		& - 	& 17.5 & -
\\ \hline	
LMT$_{c=100}$		&	3.4  & 65.4 & 92.0		 & - 	& 7.66 & -
\\ \hline	
LMT$_{c=250}$		&  6.92  & 61.8 & 98.4 		& - 	& 6.92 & -
\\ \hline	
LMT$_{c=1000}$		&  12.23 & 78.4 & 98.9 		& -	 	& 3.10 & -
\\ \hline		
Lip-NN  &  3.55 & - 	& - 		& 8.74 & - & -
\\ \hline					
\end{tabular}
\caption{Results from MNIST experiments. \label{tab:MNIST_results}}
\end{table}

%\footnotetext{Values for Lip-NN as reported in \cite{pauli2020training}. Note that their method uses a down-sampling (pooling) layer due to the computational complexity of solving a semi-definite program at each Epoch.}

\begin{table}
	\caption{Performance of convolutional LBEN versus convolutional MON on SVHN dataset. \label{tab:SVHN_results}}
	\begin{tabular}{|c||c|c|c|c|c|c|}\hline
		Model& Err:  $\|a\|_2 = 0$ & Err:  $\|a\|_2 \le 5$ &Err:  $\|a\|_2 \le  10$ & $\gamma_{up}$ & $\gamma_{low}$ & $\gamma$ approx
		\\ \hline\hline
		MON & 19.5 & 40  & 72 & - 	& 8.3 & -
		\\ \hline			
		LBEN$_{\gamma<2}$ & 22.75 & 37  & 56 & 2 	& 0.8 & 40\%
		\\ \hline			
	\end{tabular}
\end{table}
\begin{figure}
	\centering
	\includegraphics[width=0.49\linewidth, trim={2cm, 0.5cm, 3cm, 1.5cm}, clip]{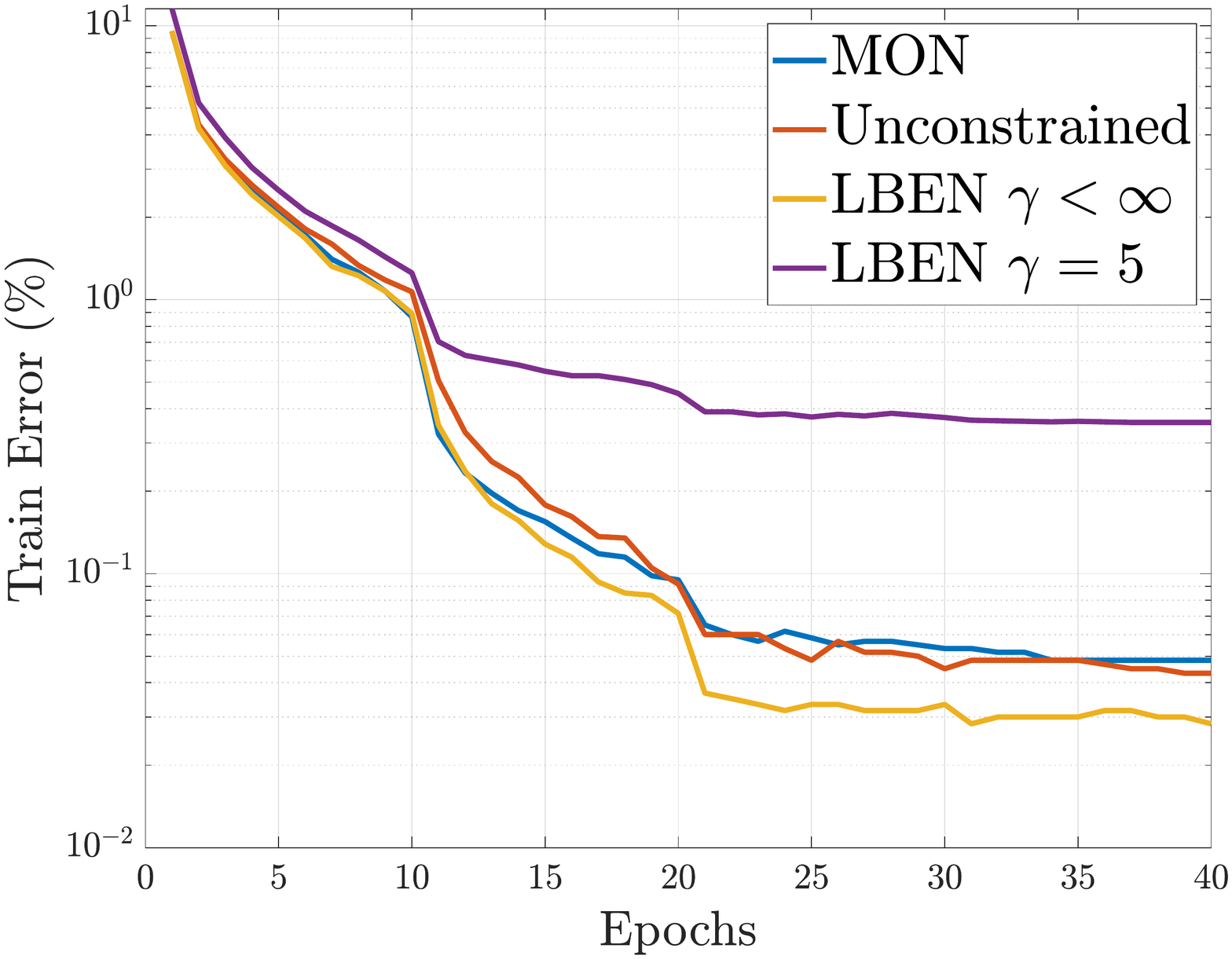} \hfill
	\includegraphics[width=0.49\linewidth, trim={2cm, 0.5cm, 3cm, 1.5cm}, clip]{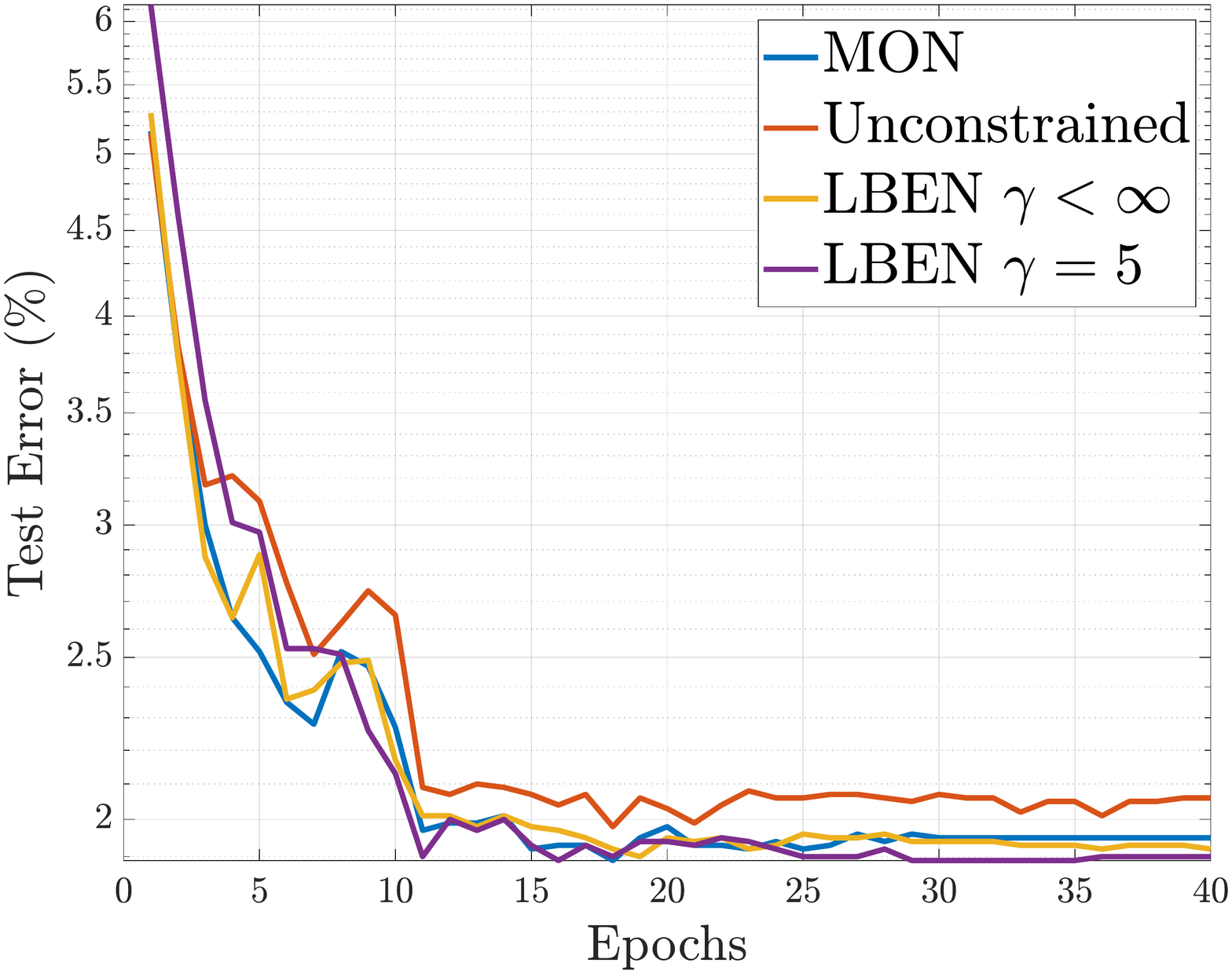}
	
	\caption{ \textbf{Left:} Training set error versus epochs. \textbf{Right:} Test set error versus epochs. Note that the left and right plots are on different scales. The time per epoch for the MON, unconstrained, LBEN$_{\gamma<\infty}$ and LBEN$_{\gamma=5}$ networks are 14.4, 16.1, 14.9 and 14.8  seconds per epoch respectively. \label{fig:mnist_training}}
\end{figure}
\newpage

\section{Monotone Operators with Non-Euclidean Inner Products}\label{sec:mono-theory}
We present some basic properties of monotone operators on a finite-dimensional Hilbert space $\Hs$, which we identify with $\R^n$ equipped with a weighted inner product $\ip{x}{y}_Q=y^\top Q x$ with $Q\succ 0$. For $n=1$, we only consider the case of $Q=1$. The induced norm $\|x\|_Q$ is defined as $\sqrt{\ip{x}{x}_Q}$. A {\em relation} or {\em operator} is a set-valued or single-valued map defined by a subset of the space $A\subseteq \Hs\times\Hs$; we use the notation $A(x)=\{y\mid (x,y)\in A\}$. If $A(x)$ is a singleton, we called $A$ a function. Some commonly used operators include: the linear operator $ A(x)=\{(x,Ax)\mid x\in\Hs\}$; the operator sum $A+B=\{(x,y+z)\mid (x,y)\in A,\, (x,z)\in B\}$; the inverse operator $A^{-1}=\{(y,x)\mid (x,y)\in A\}$; and the subdifferential operator $\partial f=\{(x,\partial f(x))\}$ with $x=\dom f$ and $\partial f(x)=\{g\in\Hs\mid f(y)\geq f(x)+\ip{y-x}{g}_Q,\,\forall y\in\Hs\}$. An operator $A$ has Lipschitz constant $L$ if for any $(x,u),(y,v)\in A$
\begin{equation}
	\|u-v\|_Q\leq L\|x-y\|_Q.
\end{equation} 
An operator $A$ is non-expansive if $L=1$ and contractive if $L<1$. An operator $A$ is monotone if 
\begin{equation}
	\ip{u-v}{x-y}_Q \geq 0, \; \forall (x,u),(y,v)\in A.
\end{equation}
It is strongly monotone with parameter $m$ if 
\begin{equation}
	\ip{u-v}{x-y}_Q \geq m\|x-y\|_Q^2, \; \forall (x,u),(y,v)\in A.
\end{equation}
A monotone operator $A$ is maximal monotone if no other monotone operator strictly contains it, which is a property required for the convergence of most fixed point iterations. Specifically, an affine operator $A(x)=Wx+b$ is (maximal) monotone if and only if $Q W+W^\top Q \succeq 0$ and strongly monotone if $ Q W+W^\top Q \succeq mI$. A subdifferential $\partial f$ is maximal monotone if and only if $f$ is a convex closed proper function. 

The resolvent and Cayley operators for an operator $A$ are denoted $R_A$ and $C_A$ and respectively defined as 
\begin{equation}
	R_A=(I+\alpha A)^{-1},\quad C_A=2R_A-I
\end{equation}
for any $\alpha>0$. When $A(x)=Wx+b$, then 
\begin{equation}
	R_A(x)=(I+\alpha W)^{-1}(x-\alpha b)
\end{equation}
and when $A=\partial f$ for some CCP function $f$, then the resolvent is given by a proximal operator
\begin{equation}
	R_A(x)=\prox_f^\alpha(x):=\argmin_{z}\frac{1}{2}\|x-z\|_Q^2+\alpha f(z).
\end{equation}
The resolvent and Cayley operators are non-expansive for any maximal monotone $A$, and are contractive for strongly monotone $A$. Operator splitting methods consider finding a zero in a sum of operators (assumed here to be maximal monotone), i.e., find $z$ such that $0\in (A+B)(z)$. For example, the convex optimization problem in (\ref{eq:cv_prob}) can be formulated as an operator splitting problem with $A(z)=(I-W)z-b$ and $B=\partial \mathfrak{f}$. Proposition~\ref{prop:operator} shows that $A$ is strongly monotone and Lipschitz with some parameters of $m$ and $L$. Here we give some popular operator splitting methods for this problem as follows.
\begin{itemize}
	\item Forward-backward splitting: $z^{k+1}=R_B(z^k-\alpha A(z^k))$, i.e.,
	\begin{equation}\label{eq:forward-backward}
		\begin{split}
			u^k&=((1-\alpha)I+\alpha W)z^k+\alpha b \\
			z^{k+1}&=\prox_{\mathfrak{f}}^\alpha (u^k)
		\end{split}
	\end{equation}
	\item Peaceman-Rachford splitting: $u^{k+1}=C_AC_B(u^k),\, z^k=R_B(u^k)$, i.e.,
	\begin{equation}\label{eq:peaceman-rachford}
		\begin{split}
			u^{k+1/2}&=2z^k-u^k, \\
			z^{k+1/2}&=(I+\alpha (I-W))^{-1}(u^{k+1/2}+\alpha b), \\
			u^{k+1}&=2x^{k+1/2}-u^{k+1/2}, \\
			z^{k+1}&=\prox_{\mathfrak{f}}^\alpha (u^{k+1}).
		\end{split}
	\end{equation}
	\item Douglas-Rachford splitting (or ADMM): $u^{k+1}=1/2(I+C_AC_B)(u^k),\, z^k=R_B(u^k)$, i.e.,
	\begin{equation}\label{eq:douglas-rachford}
		\begin{split}
			u^{k+1/2}&=2z^k-u^k, \\
			z^{k+1/2}&=(I+\alpha (I-W))^{-1}(u^{k+1/2}+\alpha b), \\
			u^{k+1}&=2x^{k+1/2}-u^{k+1/2}, \\
			z^{k+1}&=\prox_{\mathfrak{f}}^\alpha (u^{k+1}).
		\end{split}
	\end{equation}
	\item Fast iterative shrinkage-thresholding algorithm (FISTA):
	\begin{equation}\label{eq:fista}
		\begin{split}
			u^k&= \argmin_{u} \mathfrak{f}(u)+\frac{L}{2}\left\|u-\frac{1}{L}\left[(L-1)z^k+Wz^k+b\right]\right\|_2^2\\
			t^{k+1}&=\frac{1+\sqrt{1+4(t^k)^2}}{2}, \\
			z^{k+1}&=u^k+\left(\frac{t^k-1}{t^{k+1}}\right)(u^k-u^{k-1}).
			\end{split}
	\end{equation}
\end{itemize} 
A sufficient condition for forward-backward splitting to converge is $\alpha <2m/L^2$. The Peacemance-Rachford and Douglas-Rachford methods converge for any $\alpha>0$, although the convergence speed will often vary substantially based upon $\alpha$. The FISTA method converges and it does not have any hyper-parameter. When the weighting $W$ is updated, Peacemance-Rachford and Douglas-Rachford splitting need to compute a matrix inverse $(I+\alpha (I-W))^{-1}$ while FISTA requires to compute the maximum singular value of $I-W$.

%In this paper, we use Peaceman-Rachford algorithm to solve the operator splitting problem with  The iteration (\ref{eq:peaceman-rachford}) then takes the form of which converges for any $\alpha >0$, 

% In this paper, we are interested in two operator splitting methods: 1) {\em forward-backward} splitting given by the iteration
% \begin{equation}\label{eq:forward-backward}
% 	z^{k+1}=R_B(z^k-\alpha A(z^k));
% \end{equation}
% and 2) Peaceman-Rachford splitting, which is given by the iteration
% \begin{equation}\label{eq:peaceman-rachford}
% 	u^{k+1}=C_AC_B(u^k), z^k=R_B(u^k).
% \end{equation}
% A sufficient condition for forward-backward to converge is that $A$ be strongly monotone and Lipschitz with parameters of $m$ and $L$ and $\alpha <2m/L^2$; for Peacemance-Rachford, the method will for any choice of $\alpha$ if $A$ is strongly monotone. In this paper, we use Peaceman-Rachford algorithm to solve the operator spliting problem with $A(z)=(I-W)z-b$ and $B=\partial f$. The iteration (\ref{eq:peaceman-rachford}) then takes the form of
% \begin{equation}\label{eq:pr-iteration}
%   \begin{split}
%     u^{k+1/2}&=2z^k-u^k \\
%     z^{k+1/2}&=(I+\alpha (I-W))^{-1}(u^{k+1/2}+\alpha b) \\
%     u^{k+1}&=2x^{k+1/2}-u^{k+1/2} \\
%     z^{k+1}&=\prox_{f}^\alpha (u^{k+1})
%   \end{split}
% \end{equation}
% which converges for any $\alpha >0$, although the convergence speed will often vary substantially based upon $\alpha$.

\section{Proof of Theorem~\ref{thm:robustness}}\label{sec:proof-thm-2}
Rearranging Eq.~(\ref{eq:lipschitz}) yields
\[
	2\Lambda-\Lambda W-W^T\Lambda\succ \frac{1}{\gamma}(W_o^TW_o+\Lambda U U^T\Lambda)\succeq 0.
\]
The well-posedness of the equilibrium network (\ref{eq:implicit}) follows by Theorem~\ref{thm:well-posedness}. To obtain the Lipschitz bound, we first apply Schur complement to (\ref{eq:lipschitz}):
\[
\begin{split}
	\begin{bmatrix}
		2\Lambda-\Lambda W-W^\top\Lambda-\frac{1}{\gamma}W_o^\top W_o & -\Lambda U \\
		-U^\top \Lambda & \gamma I
	\end{bmatrix}\succ 0.
\end{split}	
\]
Left-multiplying $\begin{bmatrix}\Delta_z^\top & \Delta_x^\top \end{bmatrix}$ and right-multiplying $\begin{bmatrix}\Delta_z^\top & \Delta_x^\top\end{bmatrix}^\top $ gives
\[
	\begin{split}
		2\Delta_z^\top \Lambda\Delta_z-2\Delta_z^\top\Lambda W\Delta_z-\frac{1}{\gamma}\Delta_z^\top W_o^\top W_o\Delta_z-2\Delta_z^\top \Lambda U\Delta_x+\gamma \|\Delta_x\|_2^2\geq 0.
	\end{split}
\]
Since (\ref{eq:affine}) implies $\Delta_v=W\Delta_z+U\Delta_x$ and $\Delta_y=W_o\Delta_z$, the above inequality is equivalent to 
\[
	\gamma \|\Delta_x\|_2^2-\frac{1}{\gamma}\|\Delta_y\|_2^2\geq 2\Delta_z^\top \Lambda \Delta_z-2\Delta_z\Lambda\Delta_v=2\ip{\Delta_v-\Delta_z}{\Delta_z}_\Lambda.
\]
Then, the Lipschitz bound of $\gamma$ for the equilibrium network (\ref{eq:implicit}) follows by (\ref{eq:incr_sector}).

\section{Proof of Proposition~\ref{prop:sigma}}\label{sec:proof-lem-1}
({\em if}): It is well-known that if $f$ is convex closed proper function, then $\prox_f^1$ is monotone and non-expansive, i.e., it is slope-restricted in $[0,1]$. Here $f$ is not necessary to be closed as $\dom f$ (i.e. the range of $\sigma$) could be open interval $(z_l,z_r)$ or half-open interval $(z_l,z_r]$ or $[z_l,z_r)$. This can be resolved by defining $\hat{f}$ as the restriction of $f$ on the closed interval $ [\hat{z}_l,\hat{z}_r]$, and then make $\hat{z}_l\rightarrow z_l$ and $\hat{z}_r\rightarrow z_r$.

({\em only if}): Assumption~\ref{ass:sigma} implies that $\sigma$ is a non-decreasing and piece-wise differentiable function on $\R$. Then, the range of $\sigma$ is an interval, denoted by  $\mathcal{Z}$. We will construct the derivative function $f^\prime$ on $\mathcal{Z}$ first and then integrate it to obtain $f$. Let $\{z_j\in \mathcal{Z}\}_{j\in \mathbb{Z}}$ be the sequence containing all points such that either $\sigma^\prime(x_{-})=0$ or $\sigma^\prime(x_+)=0$ for all $x\in \sigma^{-1}(z_j)$. Note that $\sigma^{-1}(z)$ is a singleton for all $z\in (z_j,z_{j+1})$, whereas $\sigma^{-1}(z_j)$ is a closed interval of the forms $(-\infty,x_r]$, $[x_l,x_r]$ or $[x_l,\infty)$. Then, we define $f^\prime$ as follows
\[
	f^\prime(z)=\begin{cases}
		\min[\sigma^{-1}(z)]-z, & \text{if }z=z_j \text{ and } \min\sigma^{-1}(z)>-\infty, \\
		\max[\sigma^{-1}(z)]-z, & \text{if }z=z_j \text{ and } \min\sigma^{-1}(z)=-\infty, \\
		\sigma^{-1}(z)-z,& \text{otherwise.} 
	\end{cases}
\]
Without loss of generality, we assume that $0\in\mathcal{Z}$ and $\sigma^{-1}(0)$ is well-defined. We define the function $f$ as follows 
\[
	f(z)=\begin{cases}
		\int_0^z f^\prime(\zeta)d\zeta+C & \text{if }z\in\mathcal{Z}, \\
		\infty & \text{otherwise,}
	\end{cases}
\]
where $C$ is an arbitrary constant. Note that $f$ is a convex function as $f^\prime$ is a piecewise differentiable function on $\mathcal{Z} $ and for those points where $x=\sigma^{-1}(z)$ is well-defined, $f^\prime$ is differentiable with $f^{\prime\prime}(z)=1/\sigma^\prime(x)-1\geq 0$ as $\sigma^\prime(x)\in (0,1]$. Finally, the definition of $f^\prime$ implies that $0\in z-\sigma^{-1}(z)+\partial f(z)$, which implies that $z=\sigma(x)$ is the unique minimizer of $1/2(z-x)^2+f(z)$. Furthermore, since $\sigma$ is well-defined, we can conclude that $f$ is bounded from below. We also provide a list of $f$ for common activation functions in Table~\ref{tab:sigma-f}. A similar list can also be found in \cite{li2019lifted}.

\begin{table}[t]
	\caption{A list of common activation functions $\sigma(x)$ and associated convex proper $f(z)$ whose proximal operator is $\sigma(x)$. For $z\notin \dom f$, we have $f(z)=\infty$. In the case of Softplus activation, $\mathrm{Li}_s(z)$ is the polylogarithm function.} \label{tab:sigma-f}
	%\IM{I think the older expression for ReLU was clearer, does not require defining $\mathbbm{1}$}}\label{tab:sigma-f}
	\begin{center}
		\begin{tabular}{|c|c|c|c|}
			\hline  &&& \\
			Activation & $\sigma(x)$ & Convex $f(z)$ & $\dom f$ \\ &&& \\ \hline &&& \\
			ReLu & $\max(x,0)$ & $0$ & $[0,\infty)$\\ &&& \\
			LeakyReLu &$\max(x,0.01x)$ & $\frac{99}{2}\min(z,0)^2$ & $\R$ \\ &&& \\
			Tanh &$ \tanh(x)$ & $\frac{1}{2}\left[\ln(1-z^2)+z\ln\left(\frac{1+z}{1-z}\right)-z^2\right]$ & $(-1,1)$ \\ &&& \\
			Sigmoid &  $1/(1+e^{-x})$ & $z\ln z+(1-z)\ln(1-z)-\frac{z^2}{2}$ & $(0,1)$\\ &&& \\
			Arctan & $\arctan(x)$ & $-\ln(|\cos z|)-\frac{z^2}{2}$ & $(-1,1)$\\ &&& \\
			Softplus & $\ln(1+e^x)$ & $ -\mathrm{Li}_2(e^z)-i\pi z-z^2/2 $ & $(0,\infty)$ \\ &&& \\ \hline
		\end{tabular}
	\end{center}
\end{table}

\section{Proof of Proposition~\ref{prop:operator}}\label{sec:proof-prop-1}

The problem (\ref{eq:cv_prob}) can be formulated as a operator splitting problem $0\in (A+B)(z)$ where $ A(z)=(I-W)(z)-(Ux+b_z)$ and $B=\partial \mathfrak{f}$. The cost function $J(z)$ in (\ref{eq:cv_prob}) is strongly convex as $A$ is strictly monotone by Condition~\ref{cond:W} and $\mathfrak{f}$ is convex. Similar to \cite{winstonMonotoneOperatorEquilibrium2020}, we prove Proposition~\ref{prop:operator} by showing that the solution of (\ref{eq:implicit}), if it exists, is an fixed point of the forward-backward iteration (\ref{eq:forward-backward}) with $\alpha=1$:
\[
	z^{k+1}=R_B(z^k-\alpha Az^k)=\prox_{\mathfrak{f}}^1(z^k-\alpha(I-W)z^k+\alpha (Ux+b_z))=\sigma(Wz^k+Ux+b_z).
	%=\prox_{\mathfrak{f}}^1(Wz^k+b)=\sigma(Wz^k+b).
\]
The last equality follows by 
\[
	\sigma(x)=\begin{bmatrix}
		\argmin_{z_1}\frac{1}{2}(z_1-x_1)^2+f(z_1) \\ \vdots \\
		\argmin_{z_n}\frac{1}{2}(z_n-x_n)^2+f(z_n)
	\end{bmatrix}=\argmin_z\frac{1}{2}\|z-x\|_\Lambda^2+\sum_{i=1}^n\lambda_i f(z_i)=\prox_{\mathfrak{f}}^1(x).
\]
Note that the necessary condition for $\sigma(\cdot)$ to be diagonal is that the weight matrix $\Lambda$ is positive diagonal.

\section{Proof of Proposition~\ref{prop:gradient}}\label{sec:proof-prop-2}

The matrix $J$ is diagonal with elements in $[0, 1]$. Decompose $\Lambda = \Pi(J+\mu I)$ for some small $\mu>0$, i.e. $\Pi = \Lambda(J+\mu I)^{-1}$, which is diagonal and positive-definite. By denoting $H = \Pi(I-W) + (I-W)^T\Pi $ we obtain the following inequality from (\ref{eq:W}):
\[
	\Pi J(I-W) + (I-W)^TJ\Pi +\mu H\succeq \epsilon I,
\]
which can be rearranged as
\[
	\Pi(I-J W) + (I-J W)^T\Pi  \succeq \epsilon I+ 2\Pi(I-J)-\mu H.
\]
Since $2\Pi(I-J)\succeq 0$, we can choose a sufficiently small $\mu$ such that
\[
	\Pi(I-J W) + (I-J W)^T\Pi  \succ 0,
\]
which further implies that $I-JW$ is strongly monotone w.r.t. $\Pi$-weighted inner product, and is therefore invertible.
% 

%Theorem~\ref{thm:well-posedness} follows by the above proposition. Since (\ref{eq:neural-ode}) is time-invariant, any solution $(v(t), w(t))$ converges to the solution  $(v(t+\tau), w(t+\tau))$ for any $\tau$, i.e. the limiting solution is an equilibrium. For any fixed $b$, the equilibrium $v^*$ is unique; otherwise, any pair of distinct equilibrium points $v_1^\star$ and $v_{\star}^1$, the two solutions $v^1(\cdot)=v_{\star}^1$ and $v_2(\cdot)=v_2^\star$ of (\ref{eq:neural-ode}) do not converge to each other, which contradicts Proposition~\ref{prop:contracting-ode}. 

%\section{Proof of Proposition~\ref{prop:contracting-ode}}
% In terms of increments $\Delta_w, \Delta_v$, the affine part has linear dynamics:
% \begin{align}
% 	\dot \Delta_v =-\Delta_v+D\Delta_w
% \end{align}
% which has a transfer function
% \[
% \frac{\Delta_v(s)}{\Delta_w(s)}=G(s) := G_0(s)D, \quad G_0(s):=\frac{1}{s+1}
% \]

% We will use the following:

%I.e.
%\[
% (\Delta_w(j\omega)-\Delta_v(j\omega))^T\Lambda\Delta_w(j\omega) \ge \mu \Delta_w(j\omega)^T\Delta_w(j\omega)
%\]

% In other words, the system $G(s)$ still satisfies the condition \eqref{eq:incr_D} for all frequencies. This can be shown using positivity-preserving properties of Zames-Falb multipliers, but in this case a proof is elementary so it is included below. 

\section{Dynamical System Theory} \label{sec:dyn-sys}
In this section, we present some concepts and results of dynamical system theory that are used in this paper. We consider a nonlinear system of the form
\begin{equation}\label{eq:system}
	\dot{z}(t)=f(z(t))
\end{equation}
where $z(t)\in\R^n$ is the state, and the function $f$ is assumed to be Lipschitz continuous. By Picard's existence theorem we have a unique a solution for any initial condition. The above system is time-invariant since $f$ is not explicitly depends on $t$. System (\ref{eq:system}) is called linear time-invariant (LTI) system if $f(z)=Az+b$ for some matrix $A\in\R^{n\times n}$ and $b\in\R^n$. The point $z_\star\in\R^n$ is call an equilibrium of (\ref{eq:system}) if $f(z_\star)=0$. 

The central concern in dynamical system theory is {\em stability}. While there are many different stability notions \citep{khalil2002nonlinear}, here we mainly focus on two of them: exponential stability and contraction w.r.t a constant metric $Q\succ 0$. System (\ref{eq:system}) is said to be locally exponentially stable at the equilibrium $z_\star$ w.r.t. to the metric $Q$ if there exist some positive constants $\alpha,\beta,\delta$ such that for any initial condition $z(0)\in \mathcal{B}_\delta(z_\star):=\{z\mid \|z-z_\star\|_Q<\delta\}$, the following condition holds:
\begin{equation}\label{eq:stability}
	\|z(t)-z_\star\|\leq \alpha \|z(0)-z_\star\|_Q e^{-\beta t},\quad \forall t>0.
\end{equation}
And it is said to be globally exponentially stable if the above condition also holds for any $\delta>0$. The exponentially stability can be verified via Lyapunov's second method, i.e., finding a Lyapunov function $V=\|z\|_P^2$ with $P\succ 0$ such that $\dot{V}(t)\leq -2\beta V(t)$ along the solutions, i.e.,
\begin{equation}
	(z-z_\star)^\top Pf(z)+f(z)^\top P(z-z_\star)+2\beta (z-z_\star)^\top P (z-z_\star)\leq 0.
\end{equation}

System (\ref{eq:system}) is said to be contracting w.r.t. the metric $Q$ if there exist some positive constants $\alpha,\beta$ such that for any pair of solutions $z_1(t)$ and $z_2(t)$, we have
\begin{equation}\label{eq:contraction}
	\|z_1(t)-z_2(t)\|_Q\leq \alpha \|z_1(0)-z_2(0)\|_Q e^{-\beta t},\quad \forall t>0.
\end{equation}
Note that contraction is a much stronger notion than global exponential stability as Condition~(\ref{eq:stability}) can be implied by Condition~(\ref{eq:contraction}) by setting $z_1=z$ and $z_2=z_\star$. However, unlike the Lyapunov analysis, contraction analysis can be done via simple local analysis which does not require any prior-knowledge about the equilibrium $z_\star$. Specifically, contraction can be established by the local exponential stability of the associated differential system defined by
\begin{equation*}
	\dot{\Delta}_z=\mathrm{D}f(z) \Delta_z
\end{equation*}
where $\Delta_z(t)$ is the infinitesimal variation between $z(t)$ and its neighborhood solutions, and $\mathrm{D}f$ is Clarke generalized Jacobian. The condition for (\ref{eq:system}) to be contracting can be represented as a state-dependent Linear Matrix Inequality (LMI) as follows
\begin{equation}
	P\mathrm{D}f(z)+\mathrm{D}f(z)^\top P+2\beta P\prec 0
\end{equation}
for some $P\succ 0$ and all $z\in\R^n$. 
For an LTI system, exponential stability and contraction are equivalent and the stability condition can be s if $A$ is Hurwitz stable (i.e. all eigenvalues of $A$ have strictly negative real part). 

For most applications, the dynamic system usually involves an external input $x(t)\in \R^m$ and an output $y(t)\in\R^p$, whose state-space representation takes the form of 
\begin{equation}\label{eq:sys-io}
	\dot{z}(t)=f(z(t),x(t)), \quad y(t)=h(z(t),x(t)).
\end{equation} 
Here we measure the robustness of the above system under input perturbation by incremental $L_2$-gain. That is, system (\ref{eq:sys-io}) has an incremental $L_2$-gain bound of $\gamma$ if for any pair of inputs $x_1(\cdot),x_2(\cdot)$ with $\int_0^T\|x_1(t)-x_2(t)\|_2^2dt<\infty$ for all $T>0$, and any initial conditions $z_1(0)$ and $z_2(0)$, the solutions of (\ref{eq:sys-io}) exists and satisfy
\begin{equation}\label{eq:l2-gain}
	\int_0^T\|y_1(t)-y_2(t)\|_2^2 \, dt\leq \gamma^2\int_0^T\|x_1(t)-x_2(t)\|_2^2\, dt+\kappa(z_1(0),z_(0))
\end{equation}
for some function $\kappa(z_1,z_2)\geq 0$ with $\kappa(z,z)=0$. Note that $\gamma$ can be viewed as a Lipschitz bound of all the mappings defined by (\ref{eq:sys-io}) with some initial condition from the input signal $x(\cdot)$ to $y(\cdot)$. For any two constant inputs $x_1,x_2$, let $z_1,z_2$ and $y_1,y_2$ be the corresponding equilibrium and steady-state output, respectively. From (\ref{eq:l2-gain}) we have 
\[
	\|y_1-y_2\|_2^2\leq \|x_1-x_2\|_2^2+\kappa(z_1,z_2)/T,
\]
which implies a Lipschitz bound of $\gamma$ as $T\rightarrow\infty$.

A particular class of nonlinear systems that have strong connections to various neural networks is the so-called Lur\'{e} system, which takes the form of
\begin{equation}
	\dot{z}(t)=Az(t)+B\phi(Cz(t))
\end{equation}
where $A,B,C$ are constant matrices with proper size, and $\phi$ is a static nonlinearity with sector bounded of $[\alpha,\beta]$: for all solution $(v,w)$ with $w=\phi(v)$
\begin{equation}\label{eq:sector-bound}
	(w-\alpha v)^\top (\beta v-w)\geq 0 
\end{equation} 
or equivalently $\begin{bmatrix}
	v \\ w
\end{bmatrix}^\top \Pi 
\begin{bmatrix}
	v \\ w
\end{bmatrix}\geq 0$ with 
\begin{equation}\label{eq:iqc-multiplier}
	\Pi=\begin{bmatrix}
		2\alpha \beta I & (\alpha+\beta) I \\ (\alpha+\beta)I & -2I
	\end{bmatrix}.
\end{equation}
This implies that the origin is an equilibrium since $\phi(0)=0$. The above system can be viewed as a feedback interconnection of a linear system 
\begin{equation}
	G:\; 
	\begin{cases}
		\dot{z}(t)=Az(t)+Bw(t) \\
		v(t) = Cz(t) 
	\end{cases}
\end{equation}
and a nonlinear memoryless component $w(t)=\phi(v(t))$. The above linear system can also be described by a transfer function $G(s)$ with $s\in\mathbb{C}$. We refer to \cite{hespanhaLinearSystemsTheory2018} for details about frequency-domain concepts and results of linear systems. The frequency-domain representation for the sector bounded condition (\ref{eq:sector-bound}) can be written as
\begin{equation}
	\begin{bmatrix}
		\hat{v}(j\omega) \\ \hat{w}(j\omega)
	\end{bmatrix}^*\Pi
	\begin{bmatrix}
		\hat{v}(j\omega) \\ \hat{w}(j\omega)
	\end{bmatrix} \geq 0\quad\forall \omega\in\R
\end{equation}
where $\hat{v}(j\omega)$ and $\hat{w}(j\omega)$ are Fourier transforms of $v$ and $w$, respectively, $(\cdot)^*$ denotes the complex conjugate. Then, the closed-loop stability of the feedback interconnection can be verified by the Integral Quadratic Constraint (IQC) theorem \citep{megretski1997system}. Although the IQC framework allows for more general dynamic multipliers, here we only focus on the simple constant multiplier defined in (\ref{eq:iqc-multiplier}).
%\IM{Do we really need all this? We actually just use the KYP lemma and the circle criterion, which leads to complete (hard) IQCs, so there is no need for $\tau$}
\begin{thm}\label{thm:iqc}
	Let $G$ be stable and $\phi$ be a static nonlinearity with sector bound of $[\alpha,\beta]$. The feedback interconnection of $G$ and $\phi$ is stable if here exists $\epsilon>0$ such that
		\begin{equation}\label{eq:iqc-cond}
			\begin{bmatrix}
				G(j\omega) \\ I
			\end{bmatrix}^*\Pi
			\begin{bmatrix}
				G(j\omega) \\ I
			\end{bmatrix}\preceq -\epsilon I,\quad \forall \omega\in\R.
		\end{equation}
\end{thm}
The Kalman-Yakubovich-Popov (KYP) lemma \citep{rantzerKalmanYakubovichPopov1996} can be applied to demonstrate the equivalence of Condition 3 in Theorem~\ref{thm:iqc} to an LMI condition. The result is stated as follows.
\begin{thm}\label{thm:kyp}
	There exists a $\epsilon>0$ such that (\ref{eq:iqc-cond}) holds if and only if there exists a  matrix $P=P^\top$ such that  
	\[
		\begin{bmatrix}
			A^\top P+PA& PB \\ B^\top P & 0
		\end{bmatrix}+
		\begin{bmatrix}
			C^\top & 0 \\ 0 & I
		\end{bmatrix} \Pi
		\begin{bmatrix}
			C & 0 \\ 0 & I
		\end{bmatrix}\prec 0.
	\]
\end{thm}
\section{Proof of Proposition~\ref{prop:contraction}}\label{sec:proof-prop-4}

From (\ref{eq:neural-ode}) the dynamics of $\Delta_v$ and $\Delta_z$ can be formulated as a feedback interconnection of a linear system $\dot{\Delta}_v=-\Delta_v+W\Delta_z$ and a static nonlinearity $\Delta_z=\sigma(v_a)-\sigma(v_b)$. The linear system can be represented by a transfer function is $G(s)=1/(s+1)W$. The nonlinear component can be rewritten as $\Delta_z=\Phi(v_a,v_b)\Delta_v$ where $\Phi$ as a diagonal matrix with each $\Phi_{ii}\in[0,1]$. For the nonlinear component $\Phi$, its input and output signals satisfies the quadratic constraint (\ref{eq:incr_sector}). For the linear system $G$, we have the following lemma.

\begin{lem}\label{lem:iqc}
  If Condition~\ref{cond:W} holds, then for all $\omega\in\{\mathbb R \cup \infty\}$ 
  \begin{equation}\label{eq:freq_cond}
	  \bm{G(j\omega)\\I}^*\bm{0 & \Lambda \\ \Lambda & -2\Lambda}\bm{G(j\omega)\\I}\prec 0.
  \end{equation}
\end{lem}

%\IM{I think if we have room it would be nice to include KYP result here, since the existence of $P$ is interesting in its own right}

The KYP Lemma (Theorem~\ref{thm:kyp}) states that (\ref{eq:freq_cond}) is equivalent to the existence of a $P=P^\top$ such that
\[
\bm{-2P &PW\\W^TP&0} + \bm{0& \Lambda\\ \Lambda & -2\Lambda }\prec 0.
\]
It is clear from the upper-left block that $P\succ 0$. The above inequality also implies 
\[
2\ip{-\Delta_v+W\Delta_z}{\Delta_v}_P\le \ip{\Delta_z-\Delta_v}{\Delta_z}_\Lambda -\epsilon (\|\Delta_z\|_2^2  +\|\Delta_v\|_2^2)\leq -\epsilon (\|\Delta_z\|_2^2  +\|\Delta_v\|_2^2)\\
\]
for some $\epsilon>0$. The contraction property of the neural ODE (\ref{eq:neural-ode} follows since
\begin{equation*}
	\ddt \|\Delta_v\|_P^2=2\ip{-\Delta_v+W\Delta_z}{\Delta_v}_P \le -\epsilon (\|\Delta_z\|_2^2  +\|\Delta_v\|_2^2)\le-2\beta\|\Delta_v\|_P^2
\end{equation*}
for some sufficiently small $\beta>0$. As a byproduct of the above inequality, we will show that the operator $-f$ with with $f(v)=-v+W\sigma(v)+Ux+b_z$ is strictly monotone w.r.t. the $P$-weighted inner product since 
\[
	\ip{-f(v_a)+f(v_b)}{v_a-v_b}_P=\ip{\Delta_v-W\Delta_z}{\Delta_v}_P\geq \beta \|\Delta_v\|_P^2.
\]

\section{Proof of Lemma~\ref{lem:iqc}}
Note that (\ref{eq:freq_cond}) is equivalent to 
\begin{equation}\label{eq:lem}
	2\Lambda - G_0(j\omega)\Lambda W-G_0(-j\omega)W^T\Lambda \succeq \mu I
\end{equation}
where $G_0(j\omega)=\frac{1}{1+j\omega}$. For some $\omega\in (\mathbb R \cup \infty)$ let $g=\Re G_0(j\omega)=\Re G_0(-j\omega)$, where $\Re$ denotes real part. It is easy to verify that $g=1/(\omega^2+1)\in [0,1]$. From (\ref{eq:W}) we have
\begin{equation*}
	2g\Lambda - g\Lambda W-gW^T\Lambda\succeq g\epsilon I
\end{equation*}
for some $\epsilon >0$. Rearranging the above inequality yields
\begin{equation*}
2\Lambda - g\Lambda W-gW^T\Lambda\succeq g\epsilon I +(1-g)2\Lambda
\end{equation*}
Now, since $g\in[0,1]$ the right-hand-side is a convex combination of two positive definite matrices: $\epsilon I$ and $2\Lambda$, therefore (\ref{eq:lem}) holds for some $\mu>0$ and all $\omega\in (\mathbb R \cup \infty)$.

\section{Training Details}\label{sec:training_details}

\subsection{MNIST Example} \label{sec:mnist_training_details}
This section contains the model structures and the details of the training procedure used for the MNIST examples. All models are trained using the ADAM optimizer \cite{kingma2015adam} with an initial learning rate of $1 \times 10^3$. All models are trained for $40$ Epochs, and the learning rate is reduced by a factor of $10$ every $10$ epochs.

The models in the MNIST example are all fully connected models with 80 hidden neurons and ReLU activations.  
For the equilibrium models, the forward and backward passes models are performed using the Peaceman-Rachford iteration scheme with $\epsilon=1$ and a tolerance of $1\times10^{-2}$. When evaluating the models, we decrease the tolerance of the spitting method to $1\times 10^{-4}$.
We use the same $\alpha$ tuning procedure as \cite{winstonMonotoneOperatorEquilibrium2020}. All models were trained using the same initial point. Note that for LBEN, this requires initializing the metric $\Lambda = I$. 

The feed-forward models trained using Lipschitz margin training were trained using the original author's code which can be found at \url{https://github.com/ytsmiling/lmt}.

\subsection{SVHN Example} \label{sec:svhn_training_details}

This section contains the model structures and the details of the training procedure used for the SVHN examples. All models are trained using the ADAM optimizer \cite{kingma2015adam} with an initial learning rate of $1 \times 10^3$. The models were trained for 5 epochs and the learning rate was reduced by a factor of 10 every 10 epochs.
Each model contains a single convolutional layer with $40$ channels and a linear output layer. To encourage quick convergence of the equilbirum network solver, we set $\epsilon=5$.

The MON was evaluated using the Peaceman-Rachford Iteration scheme. 

%The implementation details of the convolutional LBEN are explained in the next section.

\subsubsection*{Convolutional LBEN}
Following the approach of \cite{winstonMonotoneOperatorEquilibrium2020}, we parametrize $U$ and $V$ in \eqref{eq:direct_gamma} via convolutions. The skew symmetric matrix is constructed by taking the skew symmetric part of a convolution $\bar{S}$, so that  $S = \frac{1}{2}(\bar{S} - \bar{S}^\top)$. 

For computational simplicity, we impose a block constant structure on the contraction metric $\Lambda = \Psi^{-1}$. In particular, if the hidden layer of the convolutional network has $n$ channels and size $s\times s$ and $W \in \mathbb{R}^{ns^2 \times n s^2}$, then we parametrize the metric as $\Psi = \bar{\Psi}\otimes I_n $ with $\bar{\Psi} \in \mathbb{R}^{s^2\times s^2}$.

In \cite{winstonMonotoneOperatorEquilibrium2020} Peaceman-Rachford is used and the operator $I-W$ can be quickly inverted using the fast Fourier transform. This situation is more complicated in our case as the term $W_\mathrm{out}^\top W_\mathrm{out}$ cannot be represented as a convolution. Instead, we apply FISTA algorithm shown in \eqref{eq:fista} which only requires the evaluation of the proximal operator. FISTA requires the calculation of the singular values of the $I-W$ which can be upper bounded via the following:
\begin{gather}
	||I-W||_2 = \left|\left|\Psi\left(\frac{1}{2\gamma}W_o^TW_o+\frac{1}{2\gamma} \Psi^{-1} U U^T \Psi^{-1}+V^TV+\epsilon I+S\right)\right|\right|_2, \\
			\leq \left|\left|\Psi\left(\frac{1}{2\gamma} \Psi^{-1} U U^T \Psi^{-1}+V^TV+\epsilon I+S\right)\right|\right|_2 + \frac{1}{2\gamma}||W_o^TW_o||_2.
\end{gather}
The first term can be quickly calculated using the approach in \cite{sedghi2018singular}. The second term can be calculated using a low rank singular value decomposition.

It should also be noted that we observed a similar trend to the \cite{winstonMonotoneOperatorEquilibrium2020}, where the Lipschitz constant of $I-W$ increases during training. This results in the number of iterations required for FISTA to converge to increase over time.

The gradient in \eqref{eq:gradient} is calculated using forward backward splitting with fixed $\alpha = 5\times 10 ^{-2}$.

\end{document}